\documentclass[10pt,twocolumn,letterpaper]{article}

\usepackage{iccv}
\usepackage{times}
\usepackage{epsfig}
\usepackage{graphicx}

\usepackage{amssymb,amsthm,amsmath,amscd}

\usepackage{epsfig}
\usepackage{subfigure}
\usepackage{algorithmic}
\usepackage{algorithm}
\usepackage{verbatim}

\theoremstyle{definition}

\newcommand{\reals}{\mathbb R}

\newcommand{\be}{\begin{equation}}
\newcommand{\ee}{\end{equation}}

\providecommand{\norm}[1]{\left\lVert#1\right\rVert}

\iccvfinalcopy 

\def\iccvPaperID{16} 

\ificcvfinal\fi
\begin{document}

\title{Kernel Spectral Curvature Clustering (KSCC)~\thanks{This work was supported by NSF grants \#0612608 and \#0915064.}}


\author{\begin{tabular}[t]{@{\extracolsep{\fill}}ccc}
Guangliang Chen & Stefan Atev & Gilad Lerman\\
School of Mathematics & Department of CS and Engineering  & School of Mathematics\\
University of Minnesota & University of Minnesota & University of Minnesota\\
127 Vincent Hall & 4-192 EE/CS Building & 127 Vincent Hall \\
206 Church Street SE & 200 Union Street SE & 206 Church Street SE \\
Minneapolis, MN 55455 & Minneapolis, MN 55455 & Minneapolis, MN 55455 \\
{\small \tt glchen@math.umn.edu} & {\small \tt atev@cs.umn.edu} & {\small \tt lerman@umn.edu}
\end{tabular}}

\maketitle

\begin{abstract}
Multi-manifold modeling is increasingly used in segmentation and
data representation tasks in computer vision and related fields.
While the general problem, modeling data by mixtures of manifolds,
is very challenging, several approaches exist for modeling data by
mixtures of affine subspaces (which is often referred to as hybrid
linear modeling). We translate some important instances of
multi-manifold modeling to hybrid linear modeling in embedded
spaces, without explicitly performing the embedding but applying the
kernel trick. The resulting algorithm, Kernel Spectral Curvature
Clustering, uses kernels at two levels - both as an implicit
embedding method to linearize nonflat manifolds and as a principled
method to convert a multi-way affinity problem into a spectral
clustering one. We demonstrate the effectiveness of the method by
comparing it with other state-of-the-art methods on both synthetic
data and a real-world problem of segmenting multiple motions from
two perspective camera views.
\end{abstract}

\noindent \textbf{Supp.~webpage}: http://www.math.umn.edu/$\sim$lerman/kscc/

\section{Introduction}
Recently a lot of attention has been focused on \emph{multi-manifold
modeling}~\cite{Agarwal05,Vidal05,Souvenir05,Kushnir06multiscale,
Yan06LSA,Ma07,Ma07Compression,Goh08Reimannian,Haro08TPMM,
spectral_theory,spectral_applied,Goldberg09ssl,Arias-Castro09Spectral}.
In a typical setting data is sampled from a mixture of distributions
approximated by manifolds (e.g., quadratic surfaces in two-view
geometries~\cite{Rao08RAS}), and the task is to segment the data
into different clusters representing the manifolds. This is a common
yet challenging problem in many applications such as computer
vision, face recognition, and image processing. A well-known example
is the clustering of the MNIST handwritten digits~\cite{LeCunMNIST},
where all the images of a given digit live on a distinct manifold.

\begin{figure*}[ht]
     \centering
     \subfigure[five circles]{\label{fig:olympics_logo}
     \includegraphics[width=.32\textwidth]{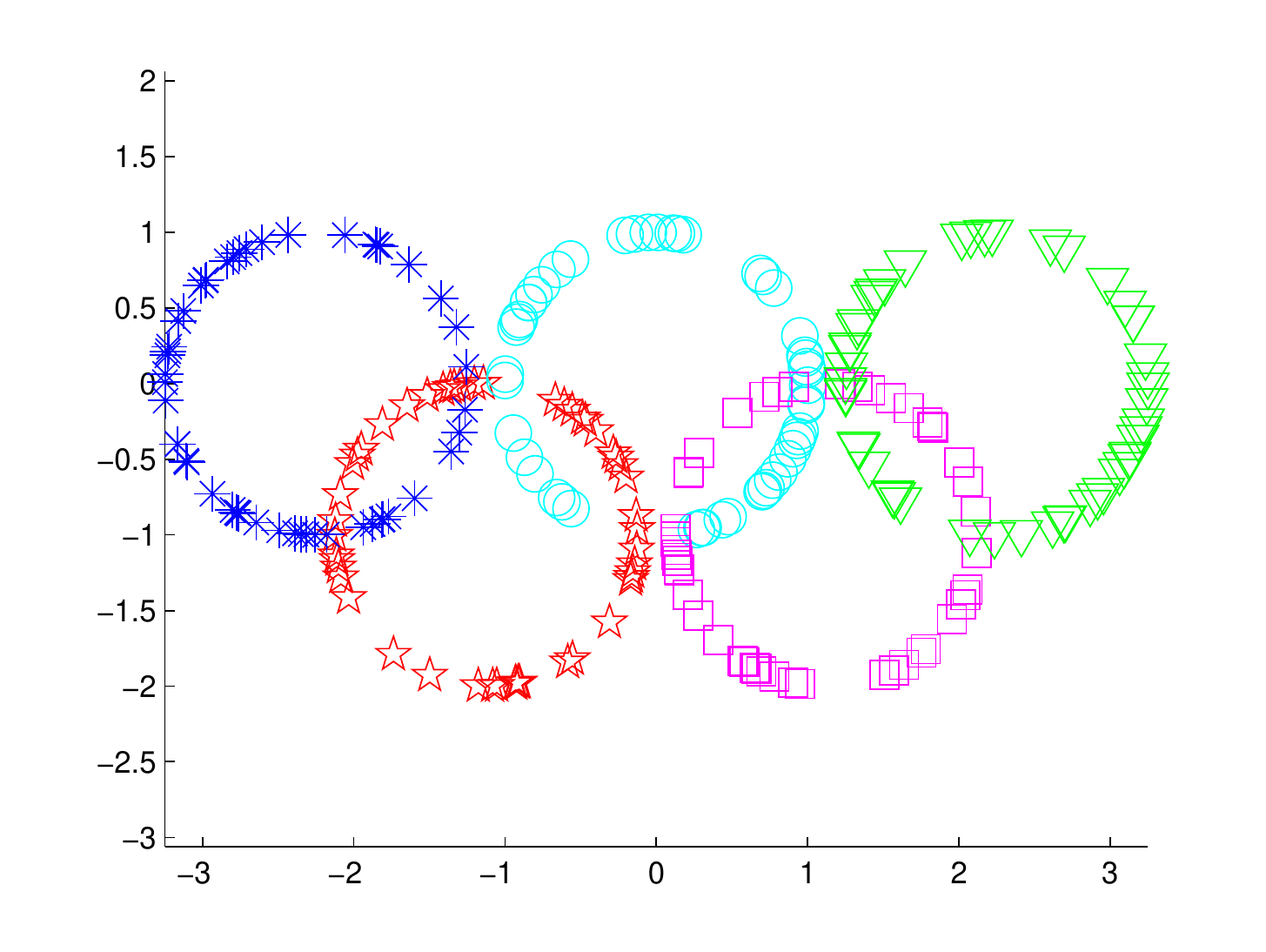}}
     \subfigure[three lines and three circles]{\label{fig:lines_circles}
     \includegraphics[width=.32\textwidth]{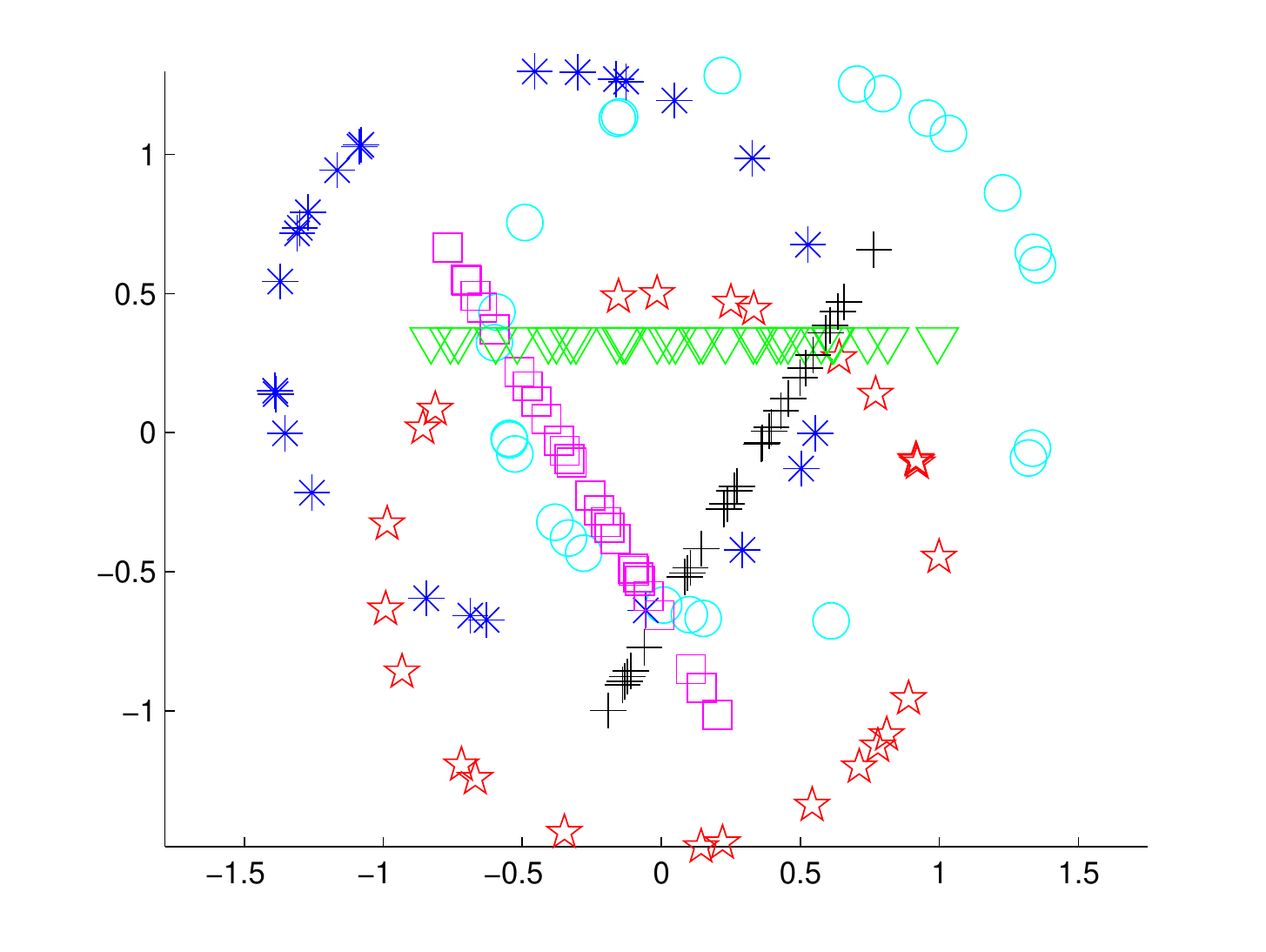}}
     \subfigure[three (noisy) spheres]{\label{fig:threespheres_4}
     \includegraphics[width=.32\textwidth]{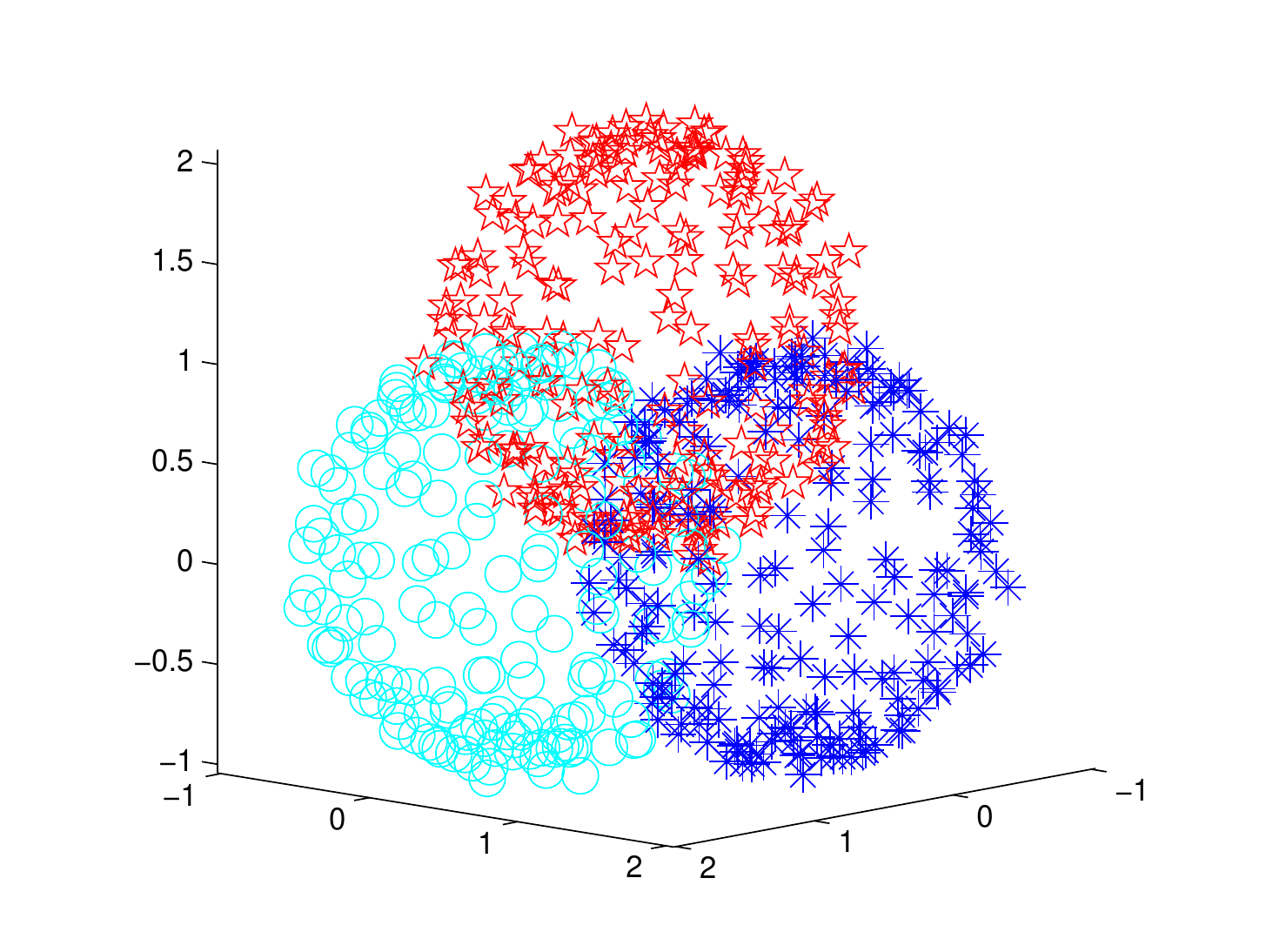}}
     \subfigure[three adjacent unit spheres and a plane through their centers]{\label{fig:plane_through_spheres}
     \includegraphics[width=.32\textwidth]{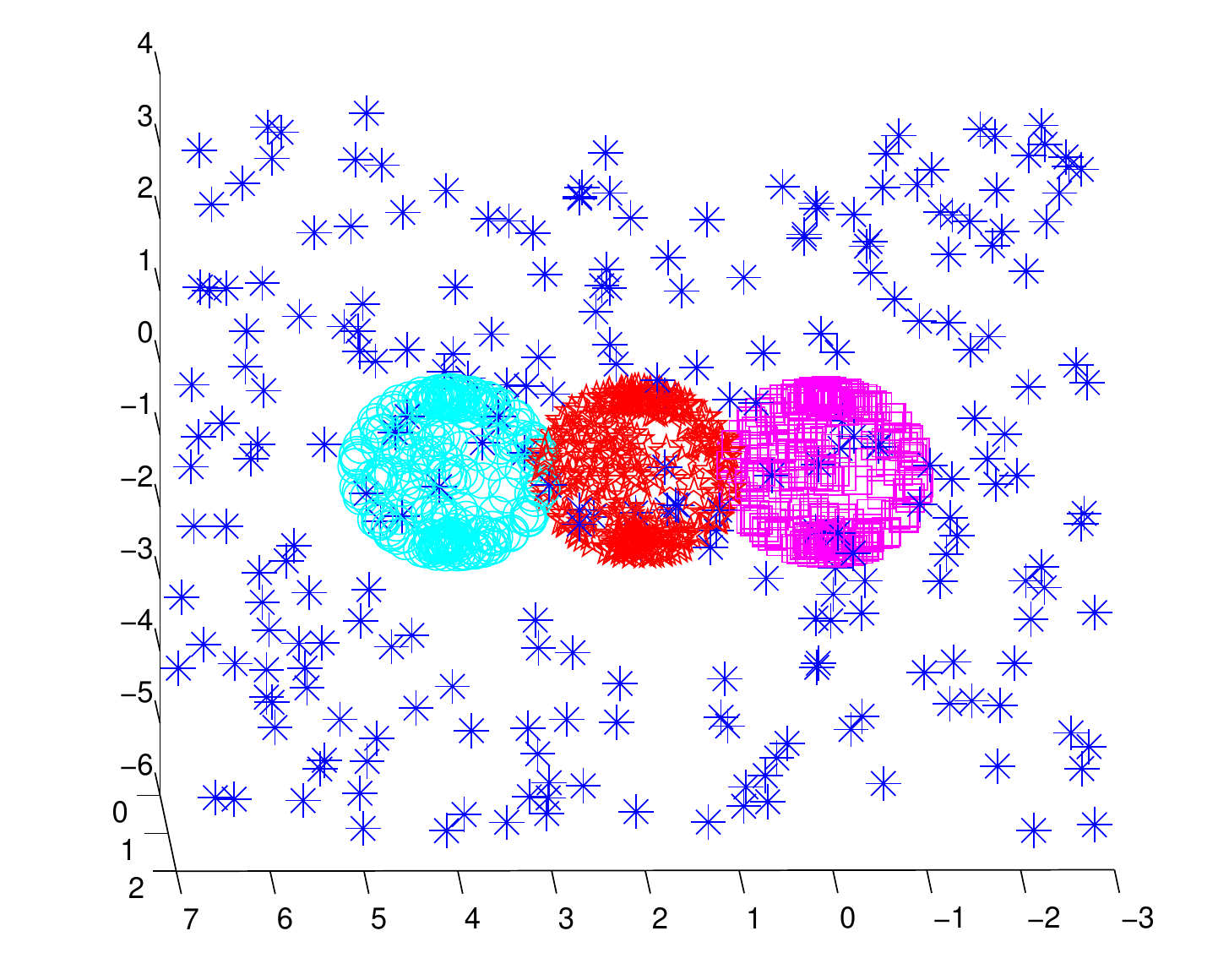}}
     \subfigure[four 1D conic sections]{\label{fig:conics_1d}
     \includegraphics[width=.32\textwidth]{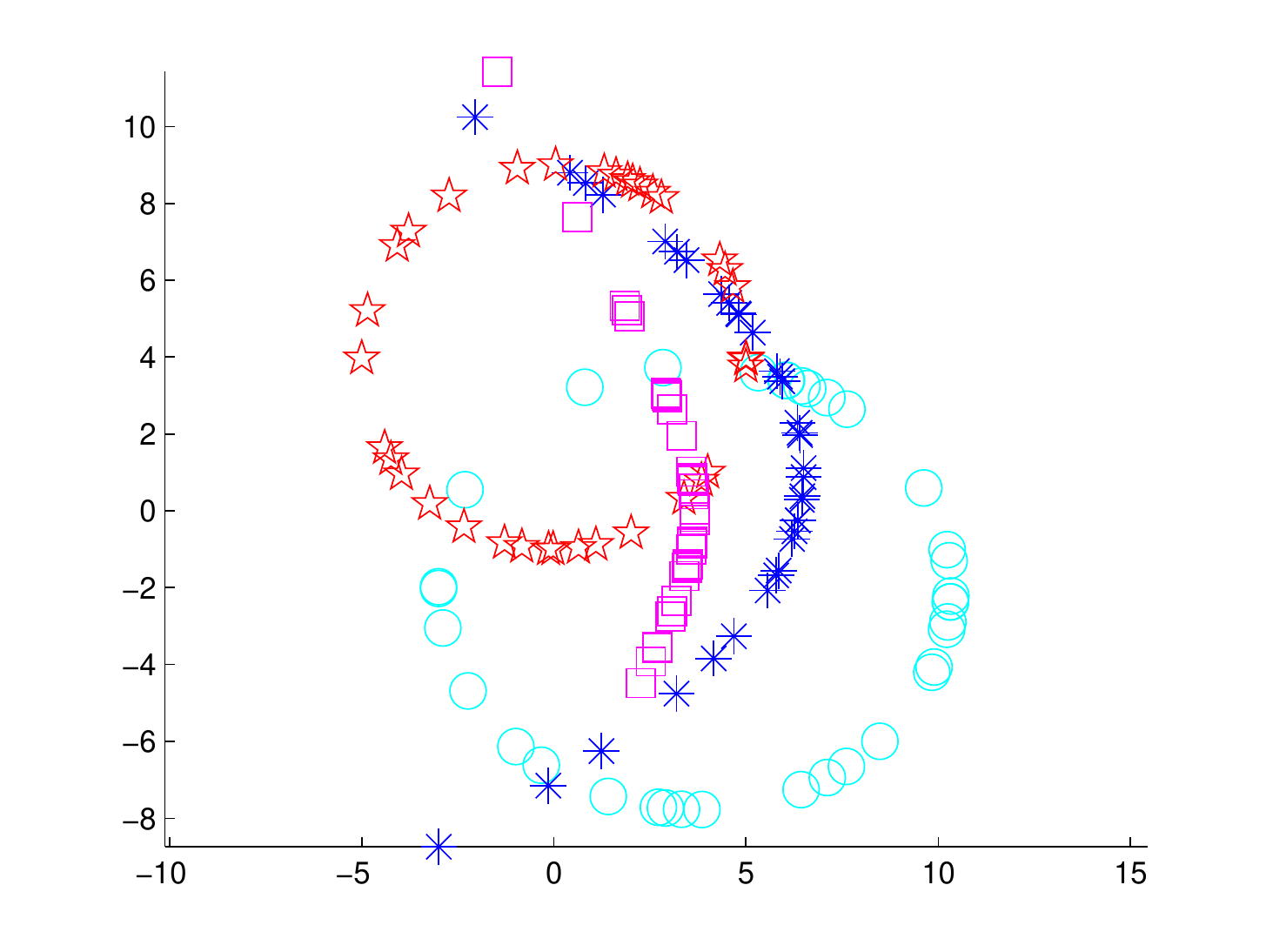}}
     \subfigure[five Lissajous curves]{\label{fig:lissajous}
     \includegraphics[width=.32\textwidth]{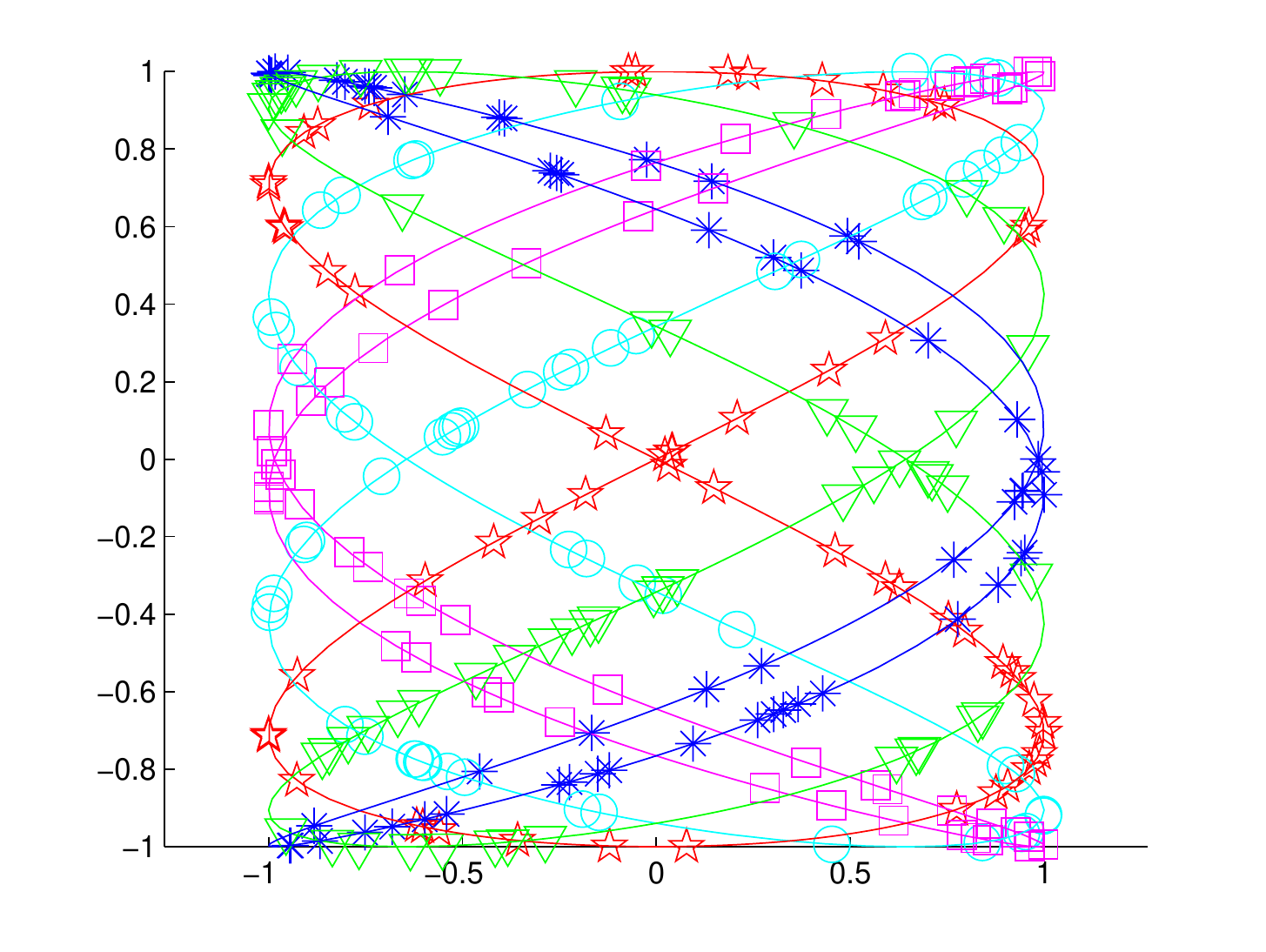}}
     \caption{Clusters obtained by the KSCC algorithm on six synthetic data sets.}
     \label{fig:artifical_data}
\end{figure*}

Due to the nature of manifolds, most algorithms analyze the local
geometry of sampled data, such as density, dimension and
orientation, and then piece together those local similarities to
find the correct
clusters~\cite{Souvenir05,Kushnir06multiscale,Goh08Reimannian,Haro08TPMM,
Goldberg09ssl,Arias-Castro09Spectral}. For example, Goldberg et
al.~\cite{Goldberg09ssl} estimate local Gaussian models around each
data point and apply \emph{spectral clustering}~\cite{Ng02}
according to the Hellinger distances between those local models. A
different local approach is used by $K$-Manifolds~\cite{Souvenir05},
which iteratively clusters data into manifolds via
expectation-maximization, i.e., first approximating each cluster by
a manifold using a node-weighted multidimensional scaling (while
using local neighbors to estimate geodesic distances), and next
assigning data points to the closest manifold from the former stage.
These methods are sensitive to a number of factors, such as size of
local neighborhoods and density of sampled data, and thus are
expected to perform poorly when data is sparsely sampled (see e.g.,
Figs.~\ref{fig:artifical_data} and \ref{fig:demo_kmanifolds}).

When only flats, i.e., affine subspaces, are used to model the
clusters, the corresponding problem, referred to as \emph{hybrid
linear modeling}, is much easier to deal with because there are
elegant representations for flats that can be utilized for solving
the problem. For example, Generalized Principal Component Analysis
(GPCA)~\cite{Vidal05,Ma07} uses polynomials to represent linear
subspaces, Local Subspace Affinity (LSA)~\cite{Yan06LSA} computes an
affinity for any pair of points using the distance between their
local tangent subspaces and then applies spectral
clustering~\cite{Ng02}, Agglomerative Lossy Compression
(ALC)~\cite{Ma07Compression} measures the number of bits needed to
code the data by general flats (up to a pre-specified distortion),
and Spectral Curvature Clustering
(SCC)~\cite{spectral_applied,spectral_theory} computes a flatness
measure for each fixed-size subset of the data. Finally, there are
algorithms that use the linear structure and iterate between a data
clustering step and a subspace estimation step, e.g.,
$K$-Flats~\cite{Kambhatla94fastnon-linear,Ho03,Bradley00kplanes,Tseng00nearest}
and Mixtures of Probabilistic PCA (MoPPCA)~\cite{Tipping99mixtures}.

\begin{figure}[t]
     \centering
     %
     \includegraphics[width=.23\textwidth]{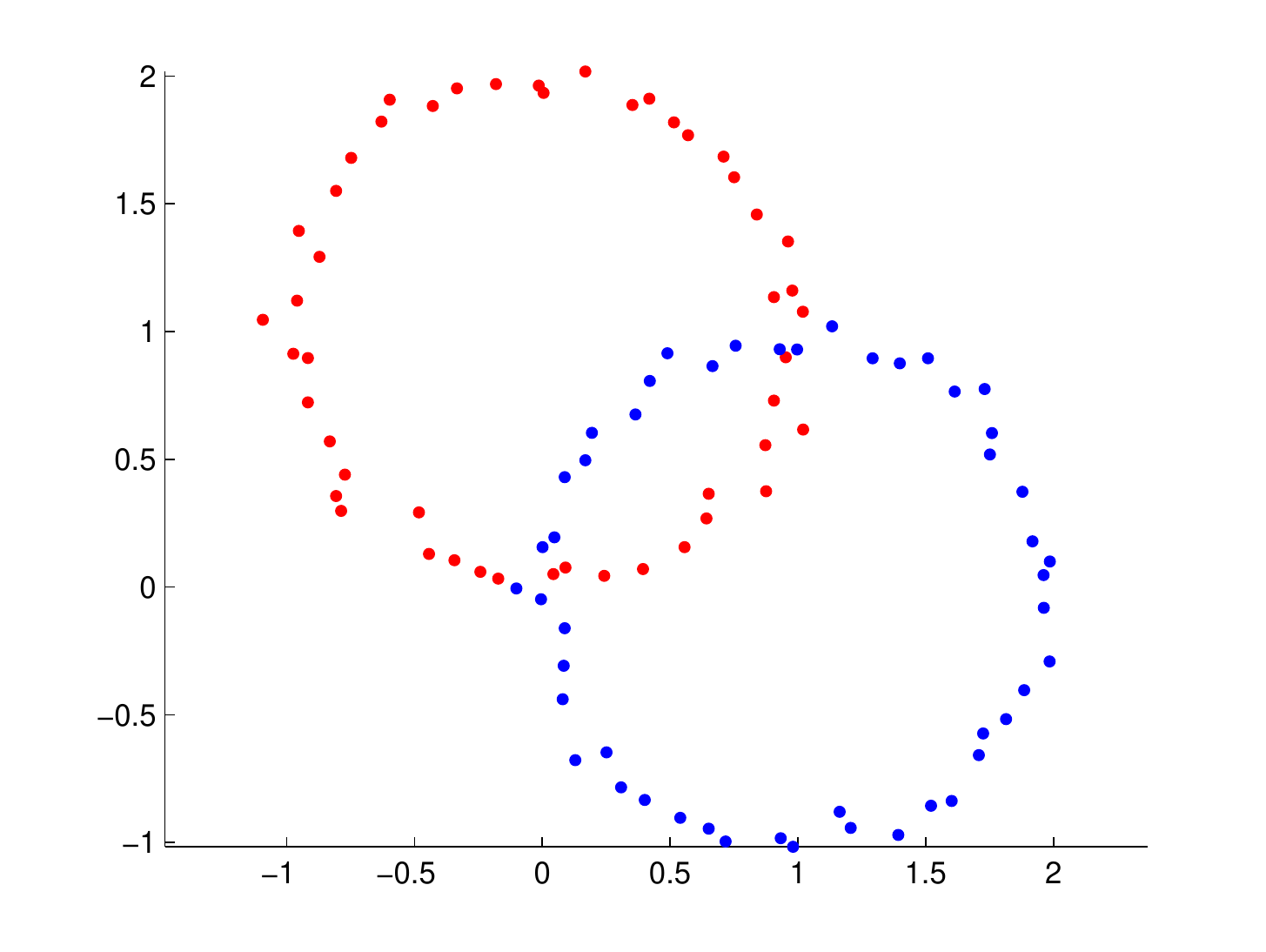}
     \includegraphics[width=.23\textwidth]{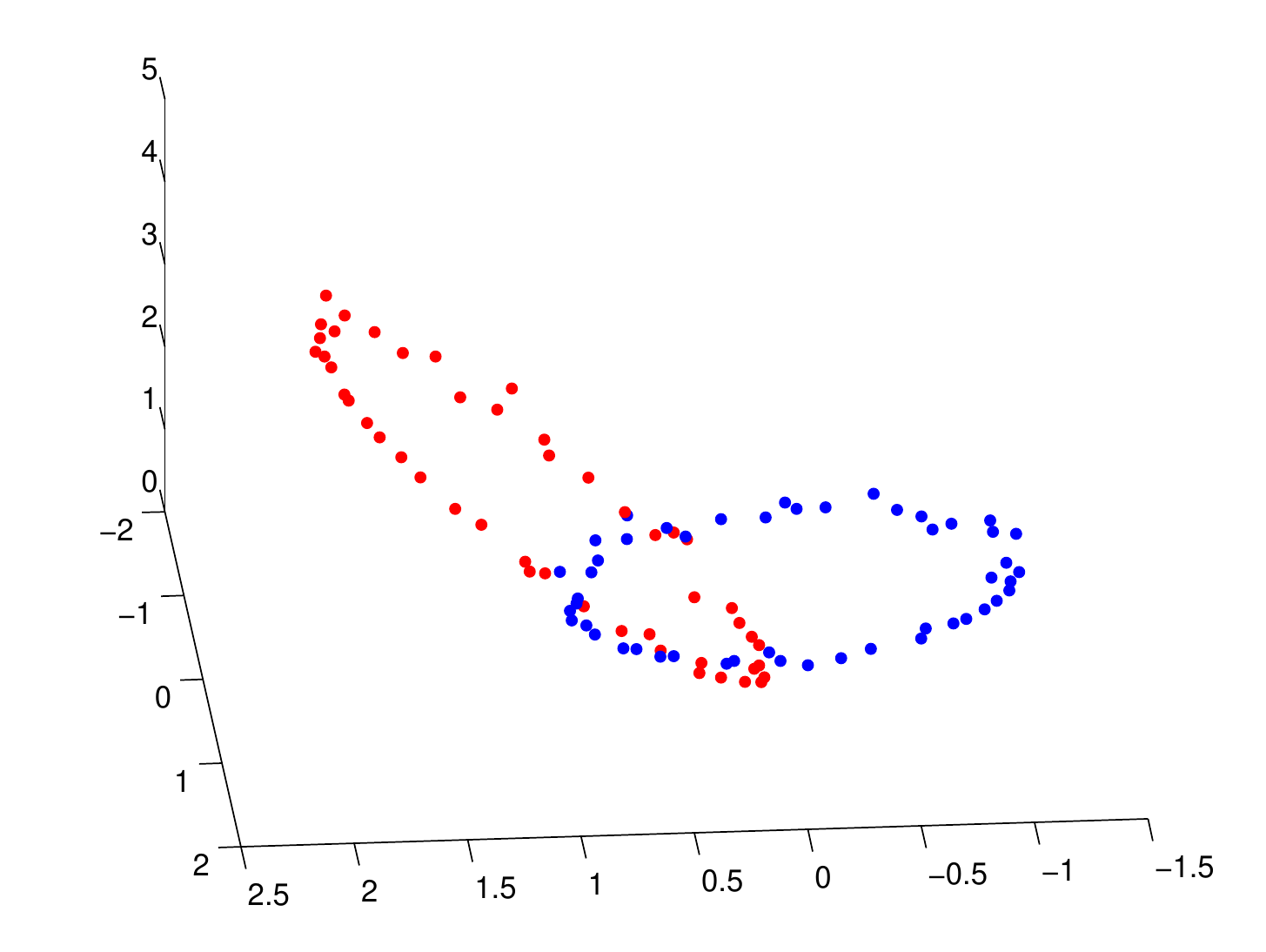}
     \caption{\small Two circles in $\mathbb{R}^2$ and their images under the map $\Phi$ (defined in Eq.~\eqref{eq:kernel_circles}) in $\mathbb{R}^3$.}
     \label{fig:circles_and_images}
\end{figure}

In this work we focus our attention on multi-manifold modeling with
parametric surfaces. Our simple but effective idea is to convert the
problem into hybrid linear modeling by embedding the underlying
(parametric) surfaces into a higher dimensional space where they
become flats. For example, when the data is sampled from a union of
$(D-1)$-dimensional hyperspheres in the Euclidean space
$\mathbb{R}^D$, the following function maps them to $D$-dimensional
flats in $\mathbb{R}^{D+1}$:
\begin{equation}\label{eq:kernel_circles}
\Phi(\mathbf{x}) = \begin{pmatrix} \mathbf{x} \\
\norm{\mathbf{x}}_2^2
\end{pmatrix}, \quad \forall\,
\mathbf{x}\in\mathbb{R}^D.
\end{equation}
Fig.~\ref{fig:circles_and_images} illustrates this example for
$D=2$. When dealing with parametric surfaces, it is possible to
apply hybrid linear modeling algorithms
(e.g.,~\cite{Bradley00kplanes,Ho03,Vidal05,Yan06LSA,Ma07Compression,
spectral_applied}) in the embedded space to segment the original
manifolds.

If a hybrid linear modeling algorithm can be expressed only in terms
of the dot products between the data points (e.g.,
$K$-Flats~\cite{Kambhatla94fastnon-linear,Ho03,
Bradley00kplanes,Tseng00nearest}, MoPPCA~\cite{Tipping99mixtures},
LSA~\cite{Yan06LSA}, ALC~\cite{Ma07Compression} and
SCC~\cite{spectral_applied,spectral_theory}), then the explicit
embedding can be avoided by using the \emph{kernel trick}. A kernel
is a real-valued function, $k(\mathbf{x},\mathbf{y})$, of two
variables $\mathbf{x},\mathbf{y} \in \mathbb{R}^D$ such that for any
$N$ points $\mathbf{x_1},\ldots,\mathbf{x_N}$ in $\reals^D$, the
\emph{kernel matrix}
\begin{equation}
\label{eq:kernel_matrix}
\mathbf{K}:=\{k(\mathbf{x_i},\mathbf{x_j})\}_{1\leq i,j\leq N}
\end{equation}
is symmetric positive semidefinite. It is shown
in~\cite{Scholkopf02Kernels} that any kernel function can be
represented as a dot product
\begin{equation}
\label{eq:kernel_by_Phi} k(\mathbf{x},\mathbf{y}) =
\langle\Phi(\mathbf{x}),\Phi(\mathbf{y})\rangle,\quad \forall\,
\mathbf{x,y} \in \mathbb{R}^D,
\end{equation}
where $\Phi\colon \mathbb{R}^D \rightarrow \mathcal{F}$ and
$\mathcal{F}$ is a Hilbert space. The map $\Phi$ is referred to as a
\emph{feature map} and the space $\mathcal{F}$ a \emph{feature
space}. Since we know the desired embedding $\Phi$, we can form the
appropriate kernel $k$ by Eq.~\eqref{eq:kernel_by_Phi} and replace
dot products with $k$ in applicable hybrid linear modeling
algorithms.

In this paper we concentrate on the kernelization of the SCC
algorithm~\cite{spectral_theory,spectral_applied}, which we refer to
as Kernel Spectral Curvature Clustering (KSCC). The main reason for
choosing SCC is that the current implementations of other hybrid
linear modeling algorithms that are appropriate for
kernelization~\cite{Tipping99mixtures,Bradley00kplanes,Ho03,Yan06LSA,
Ma07Compression} do not perform sufficiently well on affine
subspaces (unlike linear subspaces); see e.g.,
Fig.~\ref{fig:demo_gpca_lsa} and \cite[Table 2]{spectral_applied}.
Another important reason is that SCC has established theoretical
guarantees~\cite{spectral_theory} and careful numerical
estimates~\cite{spectral_applied} which can be used to justify
successes and failures of KSCC.

The rest of this paper is organized as follows. We first present the
KSCC algorithm in Section~\ref{sec:algorithm}. Experiments are then
conducted in Section~\ref{sec:experiments} to test the algorithm on
both artificial data and a real-world problem of two-view motion
segmentation (The Matlab codes and relevant data can be found at the supplemental webpage). Finally, Section~\ref{sec:conclusions} concludes with
a brief discussion and open directions for future work.

\section{The KSCC algorithm}\label{sec:algorithm}
Briefly speaking, the KSCC algorithm is the SCC
algorithm~\cite{spectral_theory,spectral_applied} performed in
some user-specified feature space. However, all the relevant
calculations in the feature space are accomplished in the original
space via the corresponding kernel function. Thus, computations
in the possibly high dimensions are avoided so as to save time.

We assume a data set
$\mathrm{X}=\{\mathbf{x}_1,\ldots,\mathbf{x}_N\}$ sampled from a
collection of $K$ manifolds in $\mathbb{R}^D$ (possibly corrupted
with noise and outliers). We will represent the data by a mixture of
$K$ parametric surfaces of the same model (e.g., general conic
sections in $\reals^2$). Based on the given model of parametric
surfaces, we form a feature map $\Phi\colon \mathbb{R}^D\rightarrow
\mathbb{R}^L$ such that the images of the $K$ parametric surfaces
are flats (see examples in Section~\ref{sec:experiments}). Let
$\ell$ be the maximal dimension of the flats. We remark that $\ell$
can be determined by subtracting 1 from the maximal number of
affinely independent coordinates of $\Phi$
(though future work will explore
substantial reduction of $\ell$, whenever possible, via a feature
selection procedure). We then segment the original manifolds by
clustering \emph{$\ell$-flats}, i.e., $\ell$-dimensional flats, in
$\mathbb{R}^L$. In practice, we use a kernel matrix $\mathbf{K}$
which is implicitly formed by the hidden embedding $\Phi$ according
to Eqs.~\eqref{eq:kernel_matrix} and~\eqref{eq:kernel_by_Phi}.

The KSCC algorithm starts by computing a \emph{polar
curvature}~\cite{spectral_applied} for any $\ell+2$ points in the
feature space via the kernel trick. Roughly speaking, the polar
curvature is an $\ell$-dimensional flatness measure (in particular,
it is zero for $\ell+2$ points lying on an $\ell$-flat). More
formally, it is the $l_2$ average of the polar sines at all vertices
of the corresponding $(\ell+1)$-simplex in the feature space,
multiplied by the diameter of that simplex.


For any set of $\ell+2$ points in the original space with indices
$\mathrm{I} =\{i_1,\ldots,i_{\ell+2}\}$, we denote the corresponding
block of the kernel matrix $\mathbf{K}$ by $\mathbf{K}_\mathrm{I,I}$
(and similarly later wherever applicable), that is,
\begin{equation} \mathbf{K}_\mathrm{I,I}:=(\mathbf{K}_{ij})_{i,j\in\mathrm{I}}.\end{equation}
The KSCC algorithm computes the polar curvature of their
corresponding feature vectors in the following way (see
supplementary material for derivation of this formula):
\begin{align} \label{eq:square_polcurv}
c^2_\mathrm{p}(\mathrm{I}) &=\frac{1}{\ell+2} \cdot
\max_{i,j\in\mathrm{I}} \left(
\mathbf{K}_{ii}+\mathbf{K}_{jj}-2\mathbf{K}_{ij} \right)
\nonumber \\
& \qquad \cdot
\sum_{i\in\mathrm{I}}\frac{\det(\mathbf{K}_\mathrm{I,I}+1)}{\prod_{j\in\mathrm{I},j\ne
i}\; \mathbf{K}_{ii}+\mathbf{K}_{jj}-2\mathbf{K}_{ij}}.
\end{align}
If any denominator above is zero then the algorithm assigns the
value 0 to the polar curvature (this happens only when two points
with indices in $\mathrm{I}$ coincide in the feature space).

The KSCC algorithm then assigns to any distinct $\ell+2$ points
(with index set $\mathrm{I}$) the following affinity:
\begin{align} \label{eq:affinity_tensor}
\mathcal{A}_\mathrm{p}(\mathrm{I}) &:=
e^{-{c^2_\mathrm{p}(\mathrm{I})/(2\sigma^2)}},
\end{align}
where $\sigma>0$ is a tuning parameter, and zero otherwise. This
function is expected to assign large values (toward 1) for points
sampled from the same parametric surface and small values (toward 0)
for points sampled from different surfaces. Its computation is
solely based on $\mathbf{K}$ without invoking directly the mapping
$\Phi$.

The KSCC algorithm next forms pairwise weights $\mathbf{W}$ from the
above multi-way affinities:
\begin{equation}\label{eq:pairwise_weights}
\mathbf{W}_{ij} = \sum_{\mathrm{J}} \mathcal{A}_\mathrm{p}(i,
\mathrm{J})\cdot \mathcal{A}_\mathrm{p}(j, \mathrm{J}),
\end{equation}
where the sum is over \begin{equation} \{\mathrm{J} = (i_1,\ldots,
i_{\ell+1}) \mid 1\leq i_1,\ldots, i_{\ell+1}\leq N\}.\end{equation}
Finally, it applies spectral clustering~\cite{Ng02} (with
$\mathbf{W}$) to find the clusters.

We have thus far described the main steps of the KSCC algorithm in
theory. However, due to its polynomial complexity ($N^{\ell+2}$),
the practical implementation of the algorithm will rely on the
numerical strategies developed in \cite{spectral_applied}, in
particular, the iterative sampling procedure for estimating the
matrix $\mathbf{W}$. This procedure is fundamental to the practical
implementation and in fact makes the KSCC a random algorithm, unlike
its brief description above. We thus provide its details below.

The zeroth iteration of the iterative sampling starts with a random
sample of $c\ll N^{\ell+1}$ $(\ell+1)$-tuples of points from the
data $\mathrm{X}$, with index sets
$\mathrm{J}_1,\ldots,\mathrm{J}_c$. It then uses them to estimate
the weights $\mathbf{W}$ of Eq.~\eqref{eq:pairwise_weights} as
follows:
\begin{equation}\label{eq:pairwise_weights_approx}
\mathbf{W}_{ij} \approx \sum_{r=1}^c \mathcal{A}_\mathrm{p}(i,
\mathrm{J}_r)\cdot \mathcal{A}_\mathrm{p}(j, \mathrm{J}_r).
\end{equation}
Based on the above weights, $K$ initial clusters are obtained by
spectral clustering~\cite{Ng02}. The first iteration then resamples
$c/K$ $(\ell+1)$-tuples of points from each of the $K$ previously
found clusters to get a better estimate of $\mathbf{W}$ so that $K$
newer (and supposedly better) clusters are found. This procedure is
repeated until convergence in order to obtain the best segmentation.
We remark that the initial sampling might be critical to good final
segmentation and, as $\ell$ becomes larger, it is increasingly
difficult to sample enough ``useful'' $(\ell+1)$-tuples of points at
the initial step.

The convergence of iterative sampling is measured by the total
\emph{kernel least squares} error, $e^2_{\text{KLS}}$, which sums
the least squares errors of $\ell$-flats approximation to the
clusters $\mathrm{C}_1,\ldots,\mathrm{C}_K$ in the feature space and
can be computed as follows (see supplementary material for proof):
\begin{align}
\label{eq:kls} e^2_{\text{KLS}}
&=\sum_{k=1}^K
\sum_{j>\ell}\lambda_j(\widetilde{\mathbf{K}}_{\mathrm{C}_k,\mathrm{C}_k}),
\end{align}
where $\lambda_j(\cdot)$ denotes the $j$-th largest eigenvalue of
the matrix, and $\widetilde{\mathbf{K}}_{\mathrm{C}_k,\mathrm{C}_k}$ is a centered version of $\mathbf{K}_{\mathrm{C}_k,\mathrm{C}_k}$ (the
block of the kernel matrix $\mathbf{K}$ corresponding to
$\mathrm{C}_k$):
\begin{align}
\widetilde{\mathbf{K}}_{\mathrm{C}_k,\mathrm{C}_k} &:=
\mathbf{K}_{\mathrm{C}_k,\mathrm{C}_k} - \mathbf{1}_{|\mathrm{C}_k|}\cdot
\mathbf{K}_{\mathrm{C}_k,\mathrm{C}_k} -
\mathbf{K}_{\mathrm{C}_k,\mathrm{C}_k} \cdot \mathbf{1}_{|\mathrm{C}_k|} \nonumber \\
& \qquad +
\mathbf{1}_{|\mathrm{C}_k|}\cdot \mathbf{K}_{\mathrm{C}_k,\mathrm{C}_k} \cdot
\mathbf{1}_{|\mathrm{C}_k|},
\end{align}
in which $|\mathrm{C}_k|$ denotes the number of points in $\mathrm{C}_k$, and $\mathbf{1}_n$ is the $n\times n$ constant matrix with elements $1/n$.

In~\cite{spectral_applied} other numerical strategies, such as an
automatic scheme of tuning the parameter $\sigma$, 
are also developed to speed
up the SCC algorithm.
We employ the same strategies to
boost the performance of the KSCC algorithm and describe the main
steps of the resulting algorithm in Algorithm~\ref{alg:kscc}.
\begin{algorithm}
\caption{Kernel Spectral Curvature Clustering (KSCC)}
\label{alg:kscc}
\begin{algorithmic}[1]
\REQUIRE Data set $\mathrm{X}$, kernel matrix $\mathbf{K}$, maximal
dimension $\ell$ (in feature space), number of manifolds $K$, and
number of sampled $(\ell+1)$-tuples $c$ (default = $100\cdot K$)
\ENSURE $K$ disjoint clusters $\mathrm{C}_1,\ldots,\mathrm{C}_K$.\\
\hspace{-.27in} \textbf{Steps:}\\
\STATE Sample randomly $c$ subsets of $\mathrm{X}$ (with indices
$\mathrm{J}_1,\ldots,\mathrm{J}_c$), each containing $\ell+1$
distinct points.
\STATE \label{step:columnwise_polcurv_computation} For each sampled
subset $\mathrm{J}_r$, compute the squared polar curvature of it and
each of the remaining $N-\ell-1$ points in $\mathrm{X}$ by
Eq.~\eqref{eq:square_polcurv}. Sort increasingly these $c\cdot
(N-\ell-1)$ squared curvatures into a vector $\mathbf{c}$.
\STATE \label{step:spectral_clustering}
\textbf{for} $p=1$ to $\ell+1$ \textbf{do}\\
\begin{itemize}
\item
Use Eq.~\eqref{eq:affinity_tensor} together with $\sigma^2 =
\mathbf{c}(N\cdot c/K^p)$ to compute the $(N-\ell-1)\cdot c$
affinities and estimate the weights $\mathbf{W}$ via
Eq.~\eqref{eq:pairwise_weights_approx}.
\item
Apply spectral clustering~\cite{Ng02} to these weights and find a
partition of the data $\mathrm{X}$ into $K$ clusters (can follow the
corresponding steps of the SCC algorithm~\cite{spectral_applied}).
\end{itemize}
\textbf{end for}\\
Record the partition $\mathrm{C}_1,\ldots,\mathrm{C}_K$ that has the
smallest total KLS error, i.e., $e_{\text{KLS}}$ of
Eq.~\eqref{eq:kls}, for the corresponding $K$ $\ell$-flats in the
feature space.
\STATE Sample $c/K$ $(\ell+1)$-tuples of points from each
$\mathrm{C}_k$ found above and repeat
Steps~\ref{step:columnwise_polcurv_computation} and
\ref{step:spectral_clustering} to find $K$ newer clusters. Iterate
until convergence to obtain a best segmentation.
\end{algorithmic}
\end{algorithm}

The KSCC algorithm employs kernels at two levels. First, it
implicitly maps each data $\mathbf{x}_i$ to a feature vector
$\mathbf{f}_i$ and uses only the kernel matrix to compute the polar
curvatures in the feature space. Second, the weight matrix
$\mathbf{W}$ (see Eq.~\eqref{eq:pairwise_weights}) can also be
interpreted as a kernel that computes dot products in the space
$\mathbb{R}^{N^{\ell+1}}$. Indeed, the feature point $\mathbf{f}_i$
is further mapped to the $i$-th slice of the tensor
$\mathcal{A}_\mathrm{p}$:
$\{\mathcal{A}_\mathrm{p}(i,i_1,\ldots,i_{\ell+1}), 1 \leq
i_1,\ldots,i_{\ell+1} \leq N\}$, which contains the interactions
between the point $\mathbf{f}_i$ and all $\ell$-flats spanned by any
$\ell+1$ points in the feature space.

\subsection{Complexity of the KSCC algorithm}
The storage requirement of the KSCC algorithm is\; $O(N\cdot(D+c))$.
The running time is $O(n_\textrm{s}\cdot (\ell+1)^2 \cdot D \cdot
N\cdot c)$, where $n_\textrm{s}$ is the number of sampling
iterations performed.

We briefly explain how to efficiently compute all the $N-\ell-1$
polar curvatures for a fixed $(\ell+1)$-tuple of points (with index
set $\mathrm{J}_r$) in
Step~\ref{step:columnwise_polcurv_computation} of
Algorithm~\ref{alg:kscc}. The complexity of computing
$\det(\mathbf{K}_{\mathrm{[i\; \mathrm{J}_r],[i\;
\mathrm{J}_r]}}+1)$ for any point $\mathbf{x}_i$ in the rest of the
data is $O((\ell+2)^3)$, which would translate into a total cost of
$O(N\cdot (\ell+2)^3)$ for all the $N-\ell-1$ curvatures. However,
in any $(\ell+2)$-tuple, $\ell+1$ of the points are the same.
Therefore, we can pre-compute all possible determinants of the form
$\mathbf{H}_{jk} =
\det(\mathbf{K}_{\mathrm{J}_r-\{j,k\},\mathrm{J}_r-\{j,k\}}+1)$ in
$O((\ell+1)^3)$ time using the fact that $\mathbf{H}=
\mathrm{adj}(\mathbf{K}_{\mathrm{J}_r,\mathrm{J}_r}+1)$. Then, each
determinant $\det(\mathbf{K}_{\mathrm{[i\; \mathrm{J}_r],[i\;
\mathrm{J}_r]}}+1)$ can be computed in $O((\ell+1)^2)$ time using
its cofactor expansion and the pre-computed minors stored in
$\mathbf{H}$, for a total cost of $O(N\cdot (\ell+1)^2)$, since
$\ell \ll N$.

\section{Numerical experiments}\label{sec:experiments}
\subsection{Artificial data}
To test the KSCC algorithm we have applied it to several artificial
data sets shown in Fig.~\ref{fig:artifical_data}.

In Figs.~\ref{fig:olympics_logo}-\ref{fig:plane_through_spheres} the
data points lie on circles/spheres and possibly also on
lines/planes. We apply the KSCC algorithm with the spherical kernel
\begin{equation} \label{eq:spherical_kernel}
k_\mathrm{s}(\mathbf{x},\mathbf{y})=\mathbf{x}'\cdot\mathbf{y}+\norm{\mathbf{x}}_2^2\cdot\norm{\mathbf{y}}_2^2,
\end{equation}
which directly follows from Eqs.~\eqref{eq:kernel_circles}
and~\eqref{eq:kernel_by_Phi}. We note that $\ell = D$ (clearly, the
$D+1$ coordinates of the mapping $\Phi$ in
Eq.~\eqref{eq:kernel_circles}, i.e.,
$\mathbf{x}_1,\ldots,\mathbf{x}_D,\norm{\mathbf{x}}^2_2$, are
affinely independent). 

In Fig.~\ref{fig:conics_1d} the data consists of a circle, an
ellipse, a parabola, and a hyperbola. It is natural to use the full
quadratic polynomial kernel
\begin{equation}\label{eq:poly2_kernel}
k_\textrm{2f}(\mathbf{x},\mathbf{y}) =
(1+\mathbf{x}'\cdot\mathbf{y})^2.
\end{equation}
This is equivalent to embedding data by the feature map
\begin{equation} \label{eq:poly2_standard_kernel}
\Phi(x_1,x_2) =
(1,\sqrt{2}x_1,\sqrt{2}x_2,x_1^2,x_2^2,\sqrt{2}x_1x_2).
\end{equation}
Therefore, the images of the 1-D conics are 4-flats in
$\mathbb{R}^6$ (the first coordinate of $\Phi$ is constant, so that
only the last five coordinates of $\Phi$ are affinely independent).
The KSCC algorithm successfully separates the different conic
sections. 


Fig.~\ref{fig:lissajous} shows five Lissajous curves in the unit
square. A Lissajous curve is the graph of the system of parametric
equations
\begin{equation}
x=A\sin(at+\delta),\quad y=B\sin(bt).
\end{equation}
We have required that $\frac{a}{b}=2$ in Fig.~\ref{fig:lissajous}.
In this case, the kernel function can be constructed as follows (see
supplementary material for proof):
\begin{align}
&k((x_1,y_1), (x_2,y_2)) \nonumber \\
& \quad = (1+ T_1(x_1)\cdot T_1(x_2)+ T_2(y_1)\cdot T_2(y_2))^2,
\end{align}
where $T_n$ is the Chebyshev polynomial of degree $n$. The KSCC
algorithm is then applied with $\ell=4$ in order to separate the
curves.

We have also tried to apply other competing algorithms to the data
in Fig.~\ref{fig:artifical_data}. Those algorithms are divided into
two categories.

\begin{figure}[t]
     \centering
     \includegraphics[width=.23\textwidth]{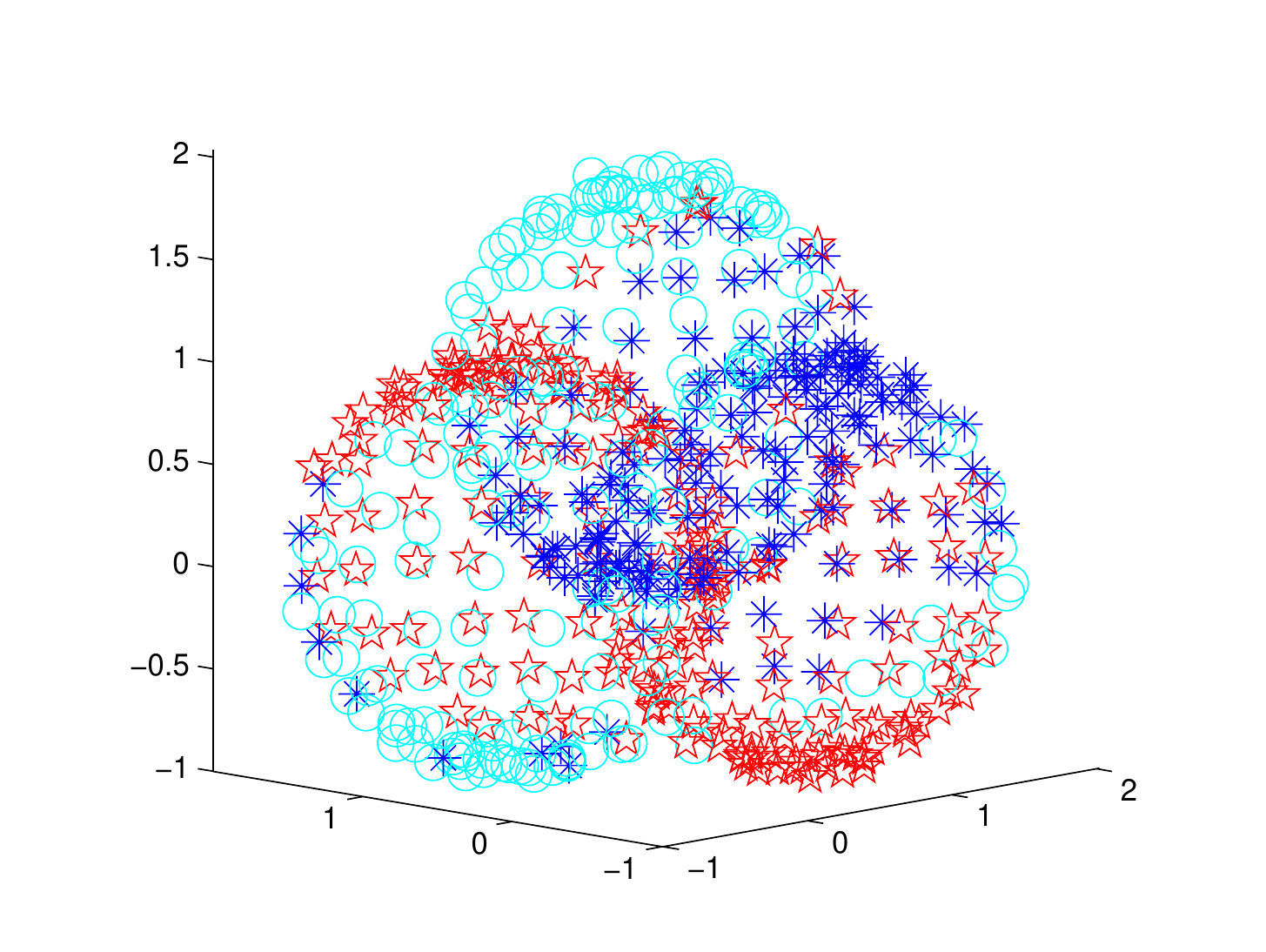}
     \includegraphics[width=.23\textwidth]{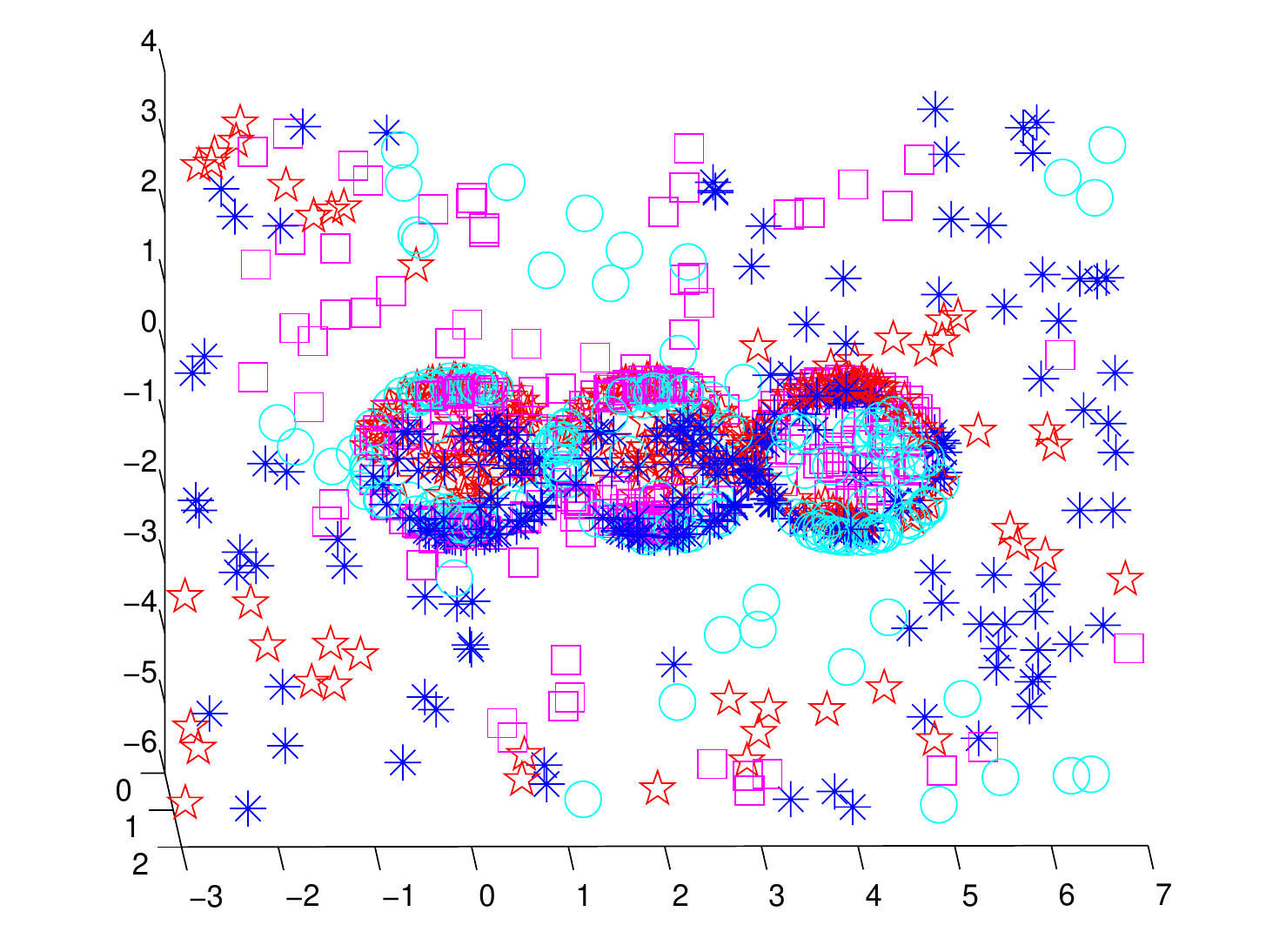}
     \caption{Demonstration of failure of local algorithms on data sets in Fig.~\ref{fig:artifical_data}.}
     \label{fig:demo_kmanifolds}
\end{figure}

The first category is local algorithms (e.g.,
\cite{Souvenir05,Kushnir06multiscale,Haro08TPMM,Goldberg09ssl,Arias-Castro09Spectral}),
i.e., algorithms that are based on local geometries. Due to the fact
that most of the data sets (in Fig.~\ref{fig:artifical_data}) are
sparsely sampled and consist of intersecting clusters, these methods
would surely fail. In addition, they are generally not suitable for
segmenting manifolds using a pre-specified model.
Fig.~\ref{fig:demo_kmanifolds} shows the failure of one such
algorithm, $K$-Manifolds~\cite{Souvenir05}, on the two most densely
sampled data sets (Figs.~\ref{fig:threespheres_4} and
\ref{fig:plane_through_spheres}). We observed in experiments that
the $K$-Manifolds algorithm tends to find arbitrary smooth manifolds
that are far from the underlying models.

\begin{figure}[t]
\centering \subfigure[GPCA (in feature space)]{\label{fig:gpca}
     \includegraphics[width=.23\textwidth]{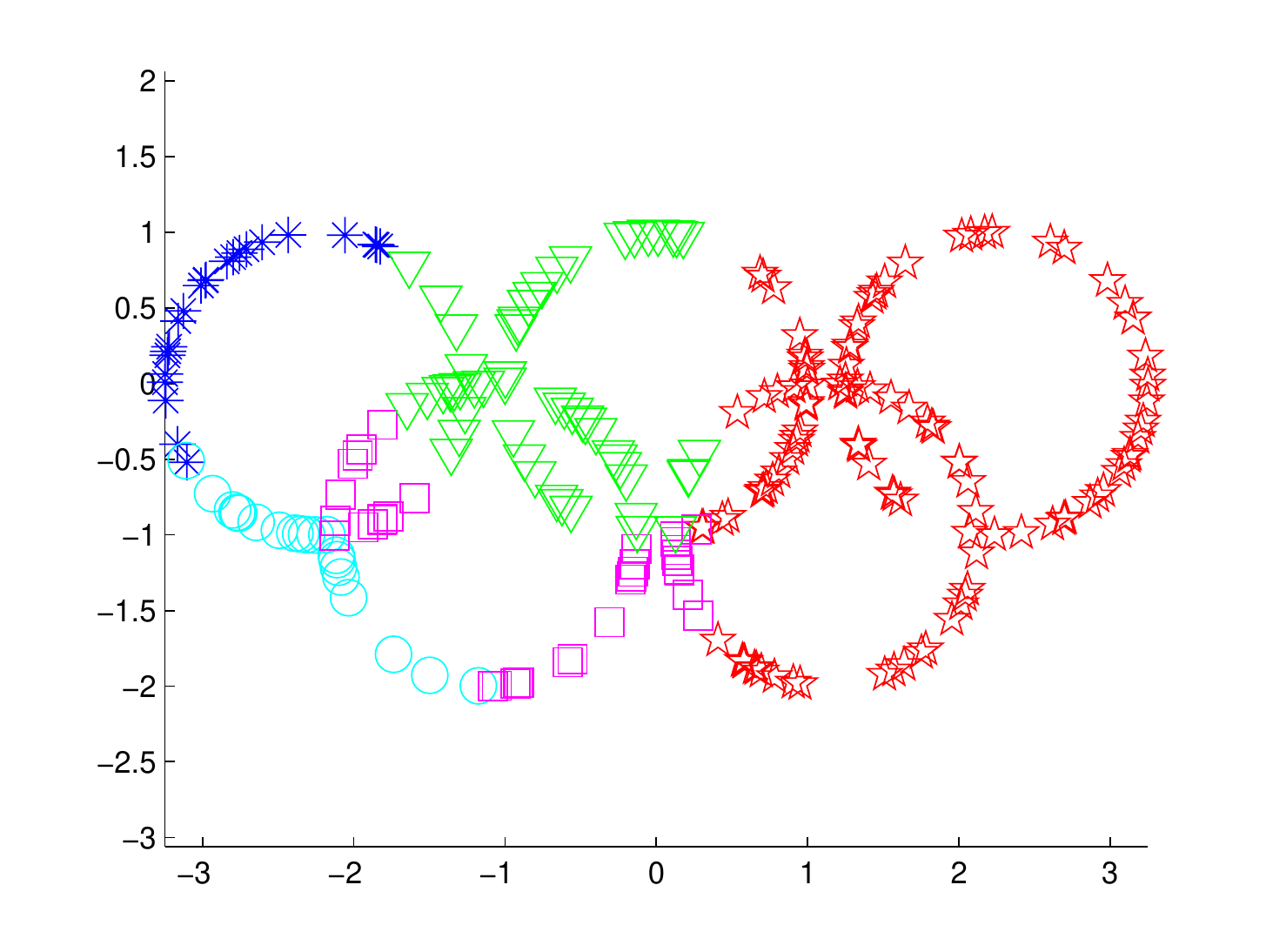}
     \includegraphics[width=.23\textwidth]{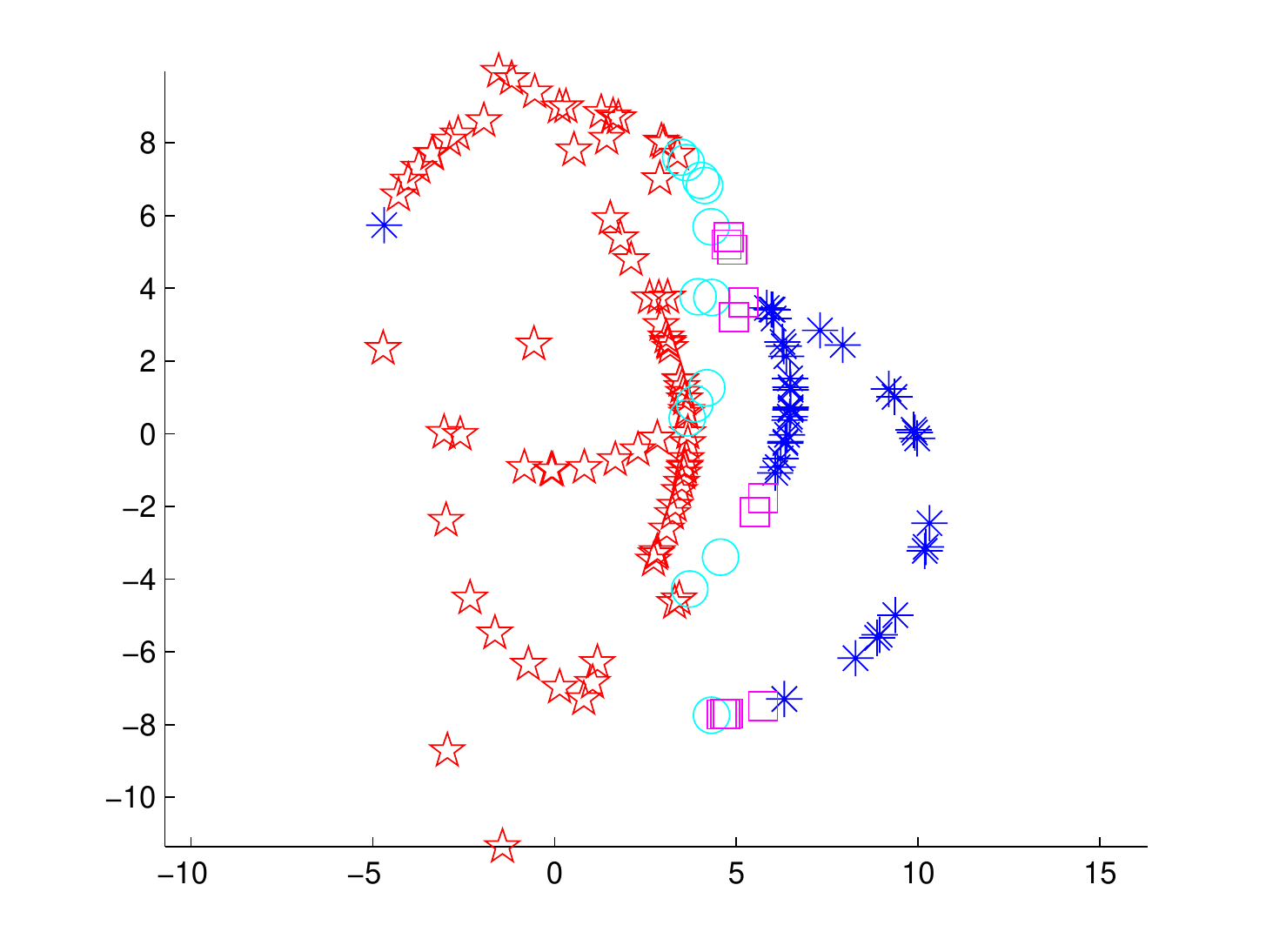}}
\subfigure[LSA (in feature space)]{\label{fig:lsa}
     \includegraphics[width=.23\textwidth]{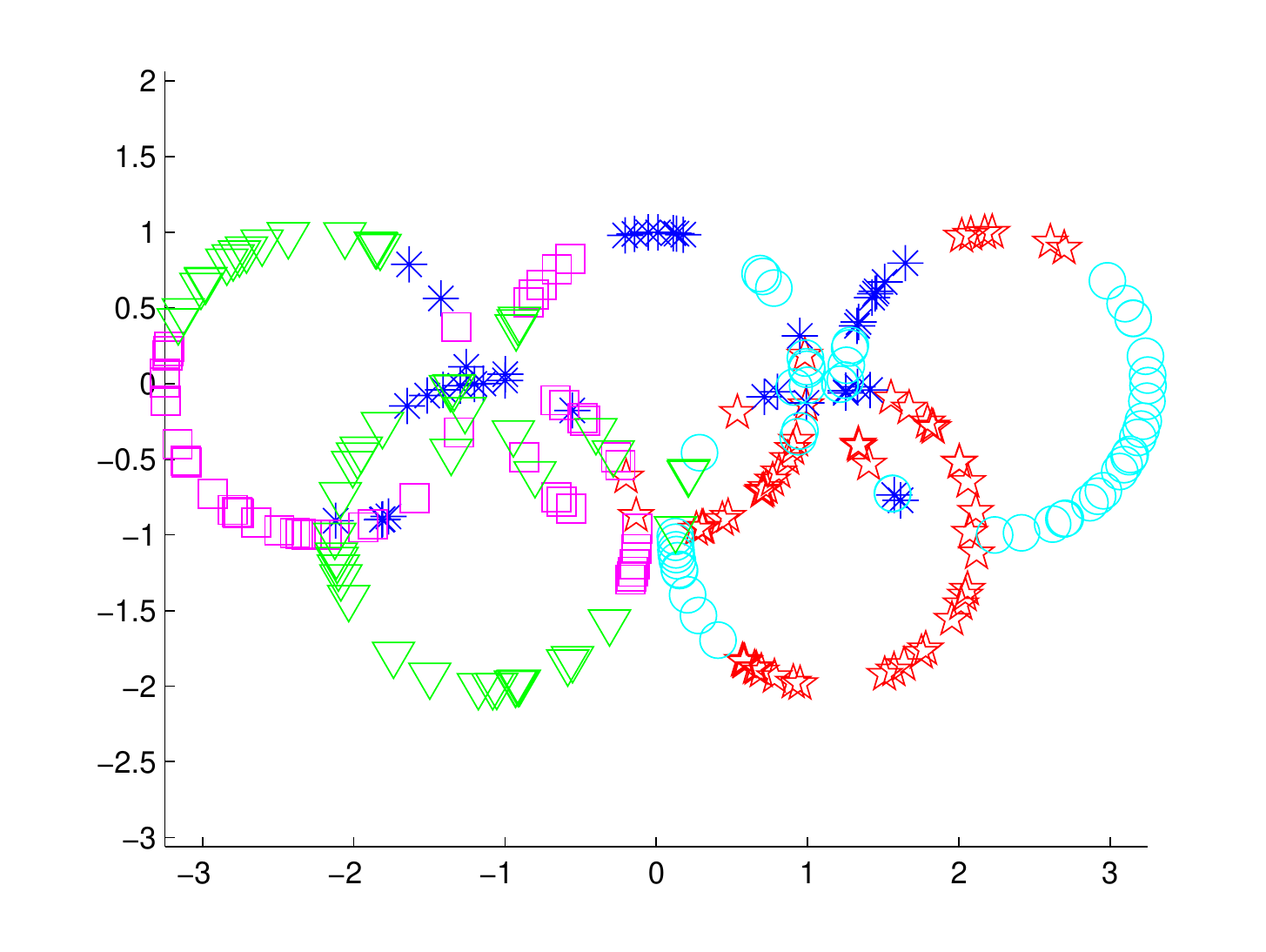}
     \includegraphics[width=.23\textwidth]{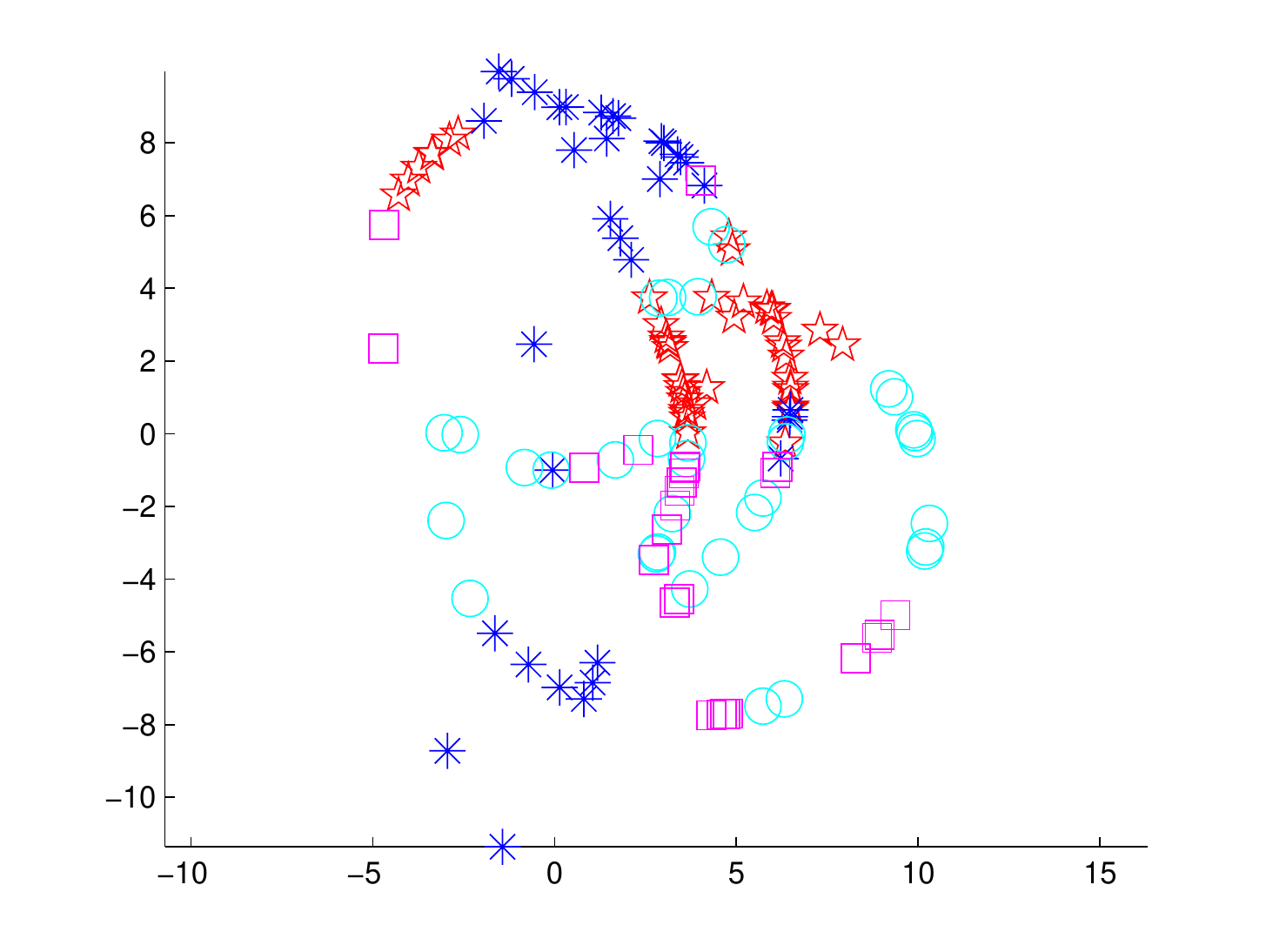}}
     \caption{Demonstration of failure of other hybrid linear
     modeling algorithms when applied to the data of Fig.~\ref{fig:artifical_data} in embedded spaces.}
\label{fig:demo_gpca_lsa}
\end{figure}

The second category is other hybrid linear modeling algorithms, such
as $K$-Flats~\cite{Kambhatla94fastnon-linear,Ho03,
Bradley00kplanes,Tseng00nearest}, MoPPCA~\cite{Tipping99mixtures},
GPCA~\cite{Vidal05,Ma07}, LSA~\cite{Yan06LSA}, and
ALC~\cite{Ma07Compression}. They can be applied to segment the
manifolds in Fig.~\ref{fig:artifical_data} in the same feature
spaces as those corresponding to KSCC, where the manifolds are
mapped to flats. However, since all these methods do not perform
well on general affine subspaces, their performance on the data in
Fig.~\ref{fig:artifical_data} (in feature space) is expected to be
very poor.

We first applied GPCA and LSA to all data sets in
Fig.~\ref{fig:artifical_data} (after being mapped to affine
subspaces), and obtained that the segmentation errors were all
around 50\%. Fig.~\ref{fig:demo_gpca_lsa} shows their segmentation
results on two data sets in Fig.~\ref{fig:artifical_data}. We also
applied ALC, $K$-Flats and MoPPCA to all the data sets (in
Fig.~\ref{fig:artifical_data}) in the feature spaces. We found that
the number of clusters found by ALC is very sensitive to its tuning
parameter ($\epsilon$), and even when ALC found the correct number
of clusters, the clusters were far from the truth. For both MoPPCA
and $K$-flats, we used ten restarts (but only recorded the best
result), and still observed that the results were all very bad. For
fair comparison, we also directly applied the SCC
algorithm~\cite{spectral_applied} in the same feature spaces and
found that it succeeded on each data set in
Fig.~\ref{fig:artifical_data}.

\subsection{Two-view motion segmentation}
\begin{figure}[t]
  \centering
  \subfigure[\textit{parking-lot}]{
     \includegraphics[width=.23\textwidth]{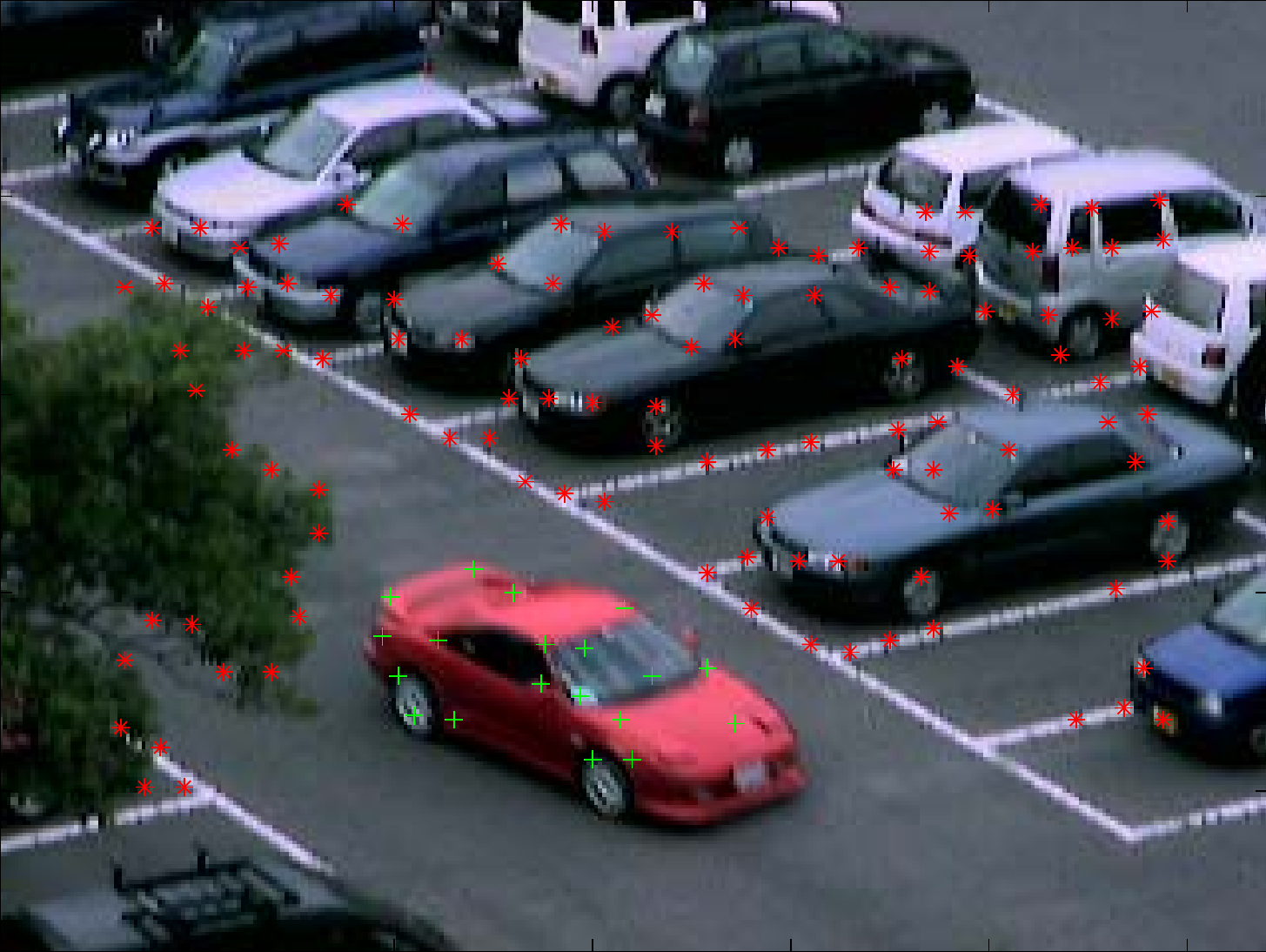}
     \includegraphics[width=.23\textwidth]{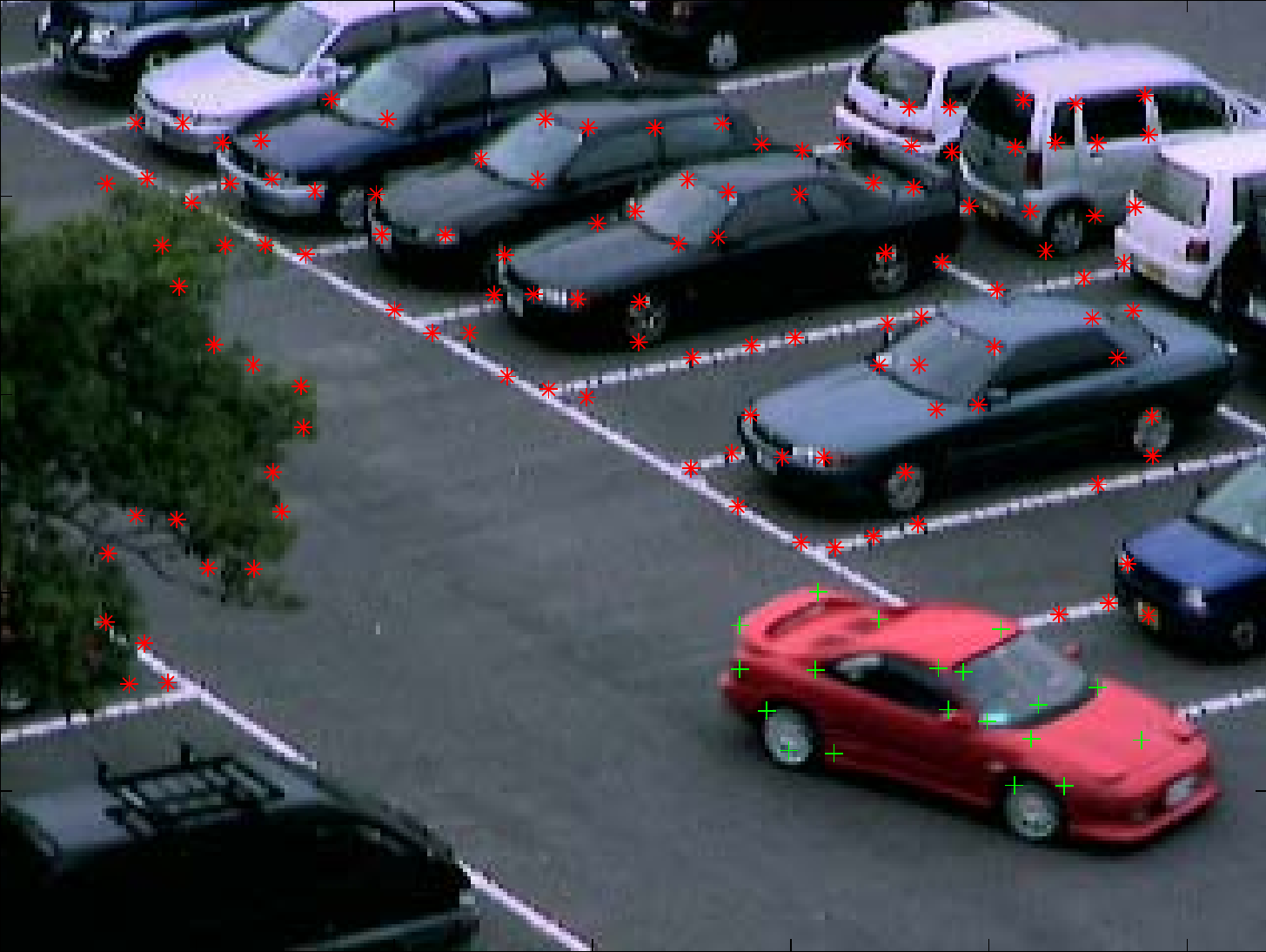}}
  \subfigure[\textit{office}]{
     \includegraphics[width=.23\textwidth]{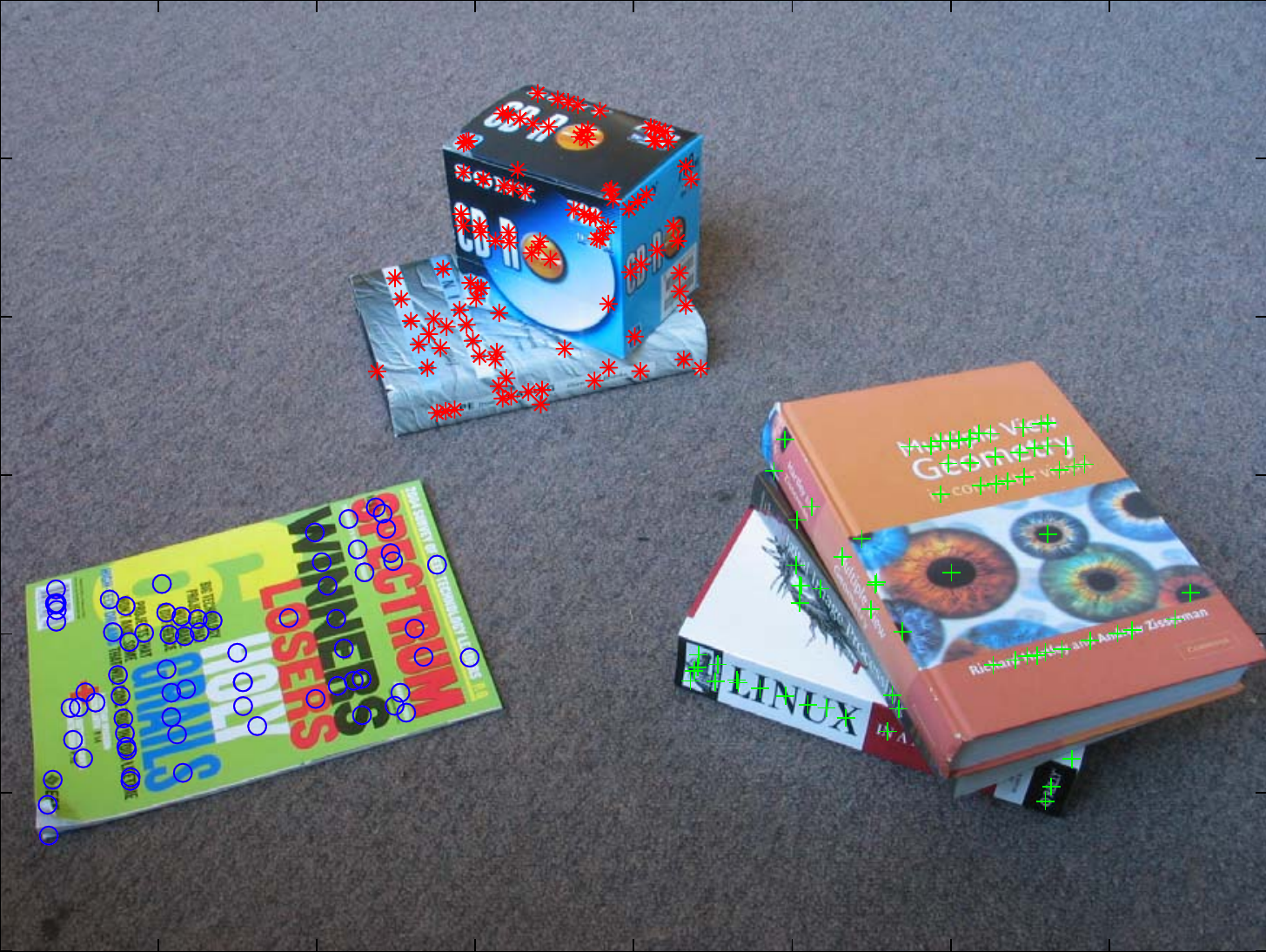}
     \includegraphics[width=.23\textwidth]{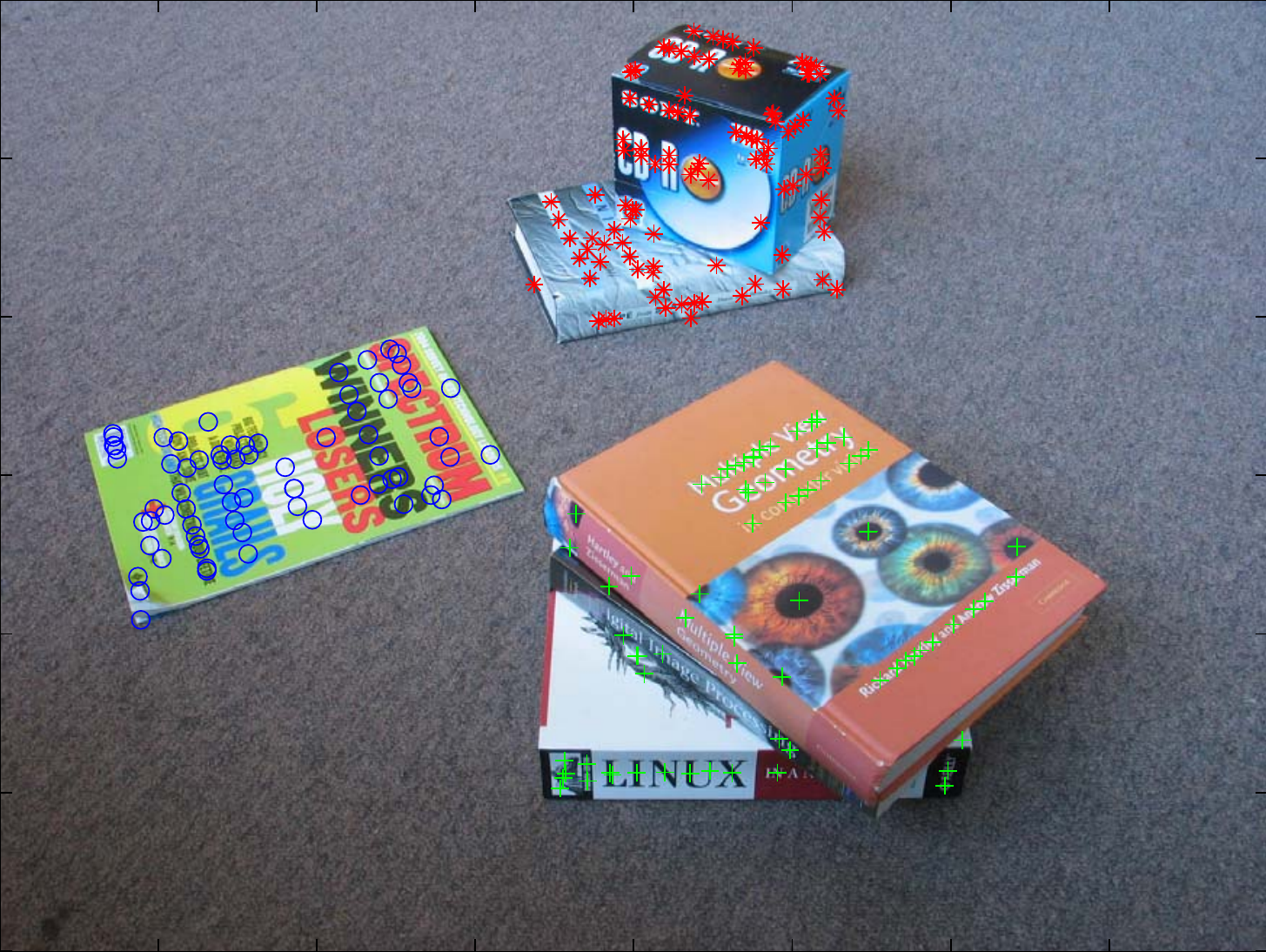}}
  \caption{Two sample sequences.}
  \label{fig:sample_seqs}
\end{figure}

In this section we compare the performance of the KSCC algorithm
with one competing method on 13 real data sequences that are studied
in~\cite{Rao08RAS} (and references therein): (1) \textit{boxes}; (2)
\textit{carsnbus3}; (3) \textit{deliveryvan}; (4) \textit{desk}; (5)
\textit{lightbulb}; (6) \textit{manycars}; (7)
\textit{man-in-office}; (8) \textit{nrbooks3}; (9) \textit{office};
(10) \textit{parking-lot}; (11) \textit{posters-checkerboard}; (12)
\textit{posters-keyboard}; and (13) \textit{toys-on-table}. Each
sequence consists of two image frames of a 3-D dynamic scene taken
by a perspective camera (see Fig.~\ref{fig:sample_seqs}), and the
task is to separate the trajectories of some feature points (tracked
on the moving objects) in the two camera views of the scene. This
application lies in the field of \emph{structure from motion}, which
is one of the fundamental problems in computer vision.

Given a point $\mathbf{x}\in \mathbb{R}^3$ in space and its image
correspondences $(x_1,y_1)', (x_2,y_2)' \in \mathbb{R}^2$ in two
views, one can form a joint image sample $\mathbf{y} =
(x_1,y_1,x_2,y_2,1)' \in \mathbb{R}^5$. It is shown (e.g.,
in~\cite{Rao08RAS}) that, under perspective camera projection, all
the joint image samples $\mathbf{y}$ corresponding to one motion
live on a distinct quadratic manifold in $\mathbb{R}^5$. More
precisely, for a 3-D rigid-body motion, there exists a symmetric
5-by-5 matrix
\begin{equation}\label{eq:H_mat}
\mathbf{H} = \begin{pmatrix} 0 & 0 & h_1 & h_2 & h_3 \\
0 & 0 & h_4 & h_5 & h_6 \\ h_1 & h_4 & 0 & 0 & h_7 \\ h_2 & h_5 & 0
& 0 & h_8 \\ h_3 & h_6 & h_7 & h_8 & h_9
\end{pmatrix}
\end{equation}
such that
\begin{equation} \label{eq:quadratic_manifold}
\mathbf{y}'\cdot \mathbf{H}\cdot \mathbf{y} = 0;
\end{equation}
and for a 2-D planar motion, there exist three matrices
$\mathbf{H}_1, \mathbf{H}_2, \mathbf{H}_3$ of the same form as in
Eq.~\eqref{eq:H_mat}, such that
\begin{equation} \label{eq:quadratic_manifold_planar}
\mathbf{y}'\cdot \mathbf{H}_i\cdot \mathbf{y} = 0, \quad i=1,2,3.
\end{equation}
This fact 
has been used by the Robust Algebraic Segmentation (RAS)
algorithm~\cite{Rao08RAS} for constructing the \emph{perspective
Veronese map} in order to segment the motions.

To solve the two-view motion segmentation problem, we also apply the
above result but will show that each motion uniquely determines a
7-flat (for 3-D rigid-body motion) or 5-flat (for 2-D planar motion)
in the space $\mathbb{R}^9$ and that KSCC can be applied to the
original 4-D point correspondences $(x_1,y_1,x_2,y_2)$ via a
properly constructed kernel function. Indeed, if we define
$\mathbf{z} := (x_1,y_1,1)\otimes (x_2,y_2,1)$, where $\otimes$
denotes the Kronecker product, then
Eq.~\eqref{eq:quadratic_manifold} can be rewritten into a linear
equation as follows:
\begin{equation}
\mathbf{z}\cdot (2h_1,2h_2,2h_3,2h_4,2h_5,2h_6,2h_7,2h_8,h_9)' = 0,
\end{equation}
and Eq.~\eqref{eq:quadratic_manifold_planar} into three such linear
equations.

Therefore, if the 4-D point correspondences $(x_1,y_1,x_2,y_2)$ are
mapped to the 9-D feature vectors $\mathbf{z}$, then one can segment
the motions by clustering 7-flats and 5-flats in $\mathbb{R}^9$ (the
last coordinate of $\mathbf{z}$ is constant). We follow the same
idea of SCC~\cite{spectral_applied} to use only the maximal
dimension when having mixed dimensions, which seems to be effective
in many cases. Therefore, we apply the KSCC algorithm with $\ell =
7$, together with the following kernel function:
\begin{align}\label{eq:twoview_kernel}
& k\left((x_1,y_1,x_2,y_2),(u_1,v_1,u_2,v_2)\right) \nonumber \\
&\quad = \left((x_1,y_1,1)\otimes
(x_2,y_2,1)\right) \cdot \left((u_1,v_1,1)\otimes (u_2,v_2,1)\right)' \nonumber \\
&\quad =(x_1u_1+y_1v_1+1)\cdot (x_2u_2+y_2v_2+1).
\end{align}

We use the outliers-free version of the 13 data sets
from~\cite{Rao08RAS} in order to solely focus on the clustering
aspect. 
We apply the KSCC algorithm (with the default $c$) to each sequence
and record the misclassification rate (in percentage) and the
running time (in seconds). To mitigate the effect of randomness due
to initial sampling, we repeat this experiment 200 times and compute
a mean error $e_\mathrm{mean}$, a standard deviation
$e_\mathrm{std}$, as well as an average running time $t$. We have
also applied the RAS algorithm~\cite{Rao08RAS} to these outlier-free
data, and found that RAS relies on an \textit{angleTolerance}
parameter (the output was quite sensitive to choices of this
parameter). We tried a few different values and combined the results
into a coherent clustering. All the experiments were performed on an
IBM T43p laptop with a 2.13 GHz Intel Pentium M processor and 1 Gb
of RAM. The results obtained by the two algorithms are summarized in
Table~\ref{tab:comparison_2view} (note that the results of RAS are
not reported in~\cite{Rao08RAS}).


\begin{table}[ht]
\centering \caption{\small The misclassification rates (in
percentage) and the running times (in seconds) of the KSCC and RAS
algorithms when applied to the 13 sequences. The second column
presents the number of samples $N$ in each sequence.
Due to randomness, the KSCC algorithm is applied 200 times to each
sequence, and a mean $e_\mathrm{mean}$ and a standard deviation
$e_\mathrm{std}$ of the errors are computed.} \vspace{.1in}
\begin{tabular}{|l|r||c|c|c||c|c|}
  \hline
    & 
    &\multicolumn{3}{c||}{KSCC}
    &\multicolumn{2}{c|}{RAS}\\
    \cline{3-7}
  Seq.  & $N$  & $e_\mathrm{mean}$ & $e_\mathrm{std}$ & $t$ & $e$ & $t$\\ 
  \hline\hline
  1  & 236 & 0.85\% & 0.00\%  & 2.14 & 0.85\% & 6.68\\
  2  & 219 & 1.04\% & 5.63\%  & 2.05 & 0.00\% & 6.68\\
  3  & 254 & 30.8\% & 5.59\%  & 2.59 & 15.4\% & 6.84\\
  4  & 155 & 0.22\% & 1.04\%  & 1.99 & 5.16\% & 4.50\\
  5  & 205 & 0.00\% & 0.00\%  & 1.86 & 0.00\% & 5.29\\
  6  & 144 & 15.7\% & 8.96\%  & 0.70 & 0.00\% & 3.91\\
  7  & 73  & 0.63\% & 2.76\%  & 1.87 & 15.1\% & 1.17\\
  8  & 388 & 1.97\% & 5.45\%  & 3.43 & 11.1\% & 20.2\\
  9  & 259 & 0.05\% & 0.13\%  & 2.18 & 0.00\% & 6.68\\
  10 & 136 & 22.3\% & 18.7\%  & 1.18 & 0.00\% & 2.89\\
  11 & 280 & 4.97\% & 1.03\%  & 2.68 & 0.00\% & 10.5\\
  12 & 297 & 1.38\% & 0.80\%  & 2.62 & 0.34\% & 9.49\\
  13 & 91  & 2.89\% & 3.78\%  & 0.87 & 18.7\% & 1.84\\
  \hline
\end{tabular}
\label{tab:comparison_2view}
\end{table}

We observe that the KSCC and RAS algorithms (a) obtain the same
classification error on sequences 1, 5, 9 (on sequence 9 the
difference is negligible: $0.05\%\cdot 259 = 0.13$); (b) have
significantly different misclassification rates (i.e., with a
difference larger than 4\%) on eight sequences (3, 4, 6, 7, 8, 10,
11, 13), with each algorithm having a better performance on four of
them (KSCC on sequences 4, 7, 8, 13, and RAS on sequences 3, 6, 10,
11); (c) have very small difference on sequences 2, 12 (about 1\%,
i.e., 3 points). In terms of running time, the KSCC algorithm is at
least twice as fast as RAS on all sequences (except sequence 7),
sometimes even being five times faster (on sequence 8). In summary,
the KSCC and RAS algorithms have almost comparable performances in
terms of segmentation errors; however, the KSCC algorithm is faster.

\section{Discussions and future work}\label{sec:conclusions}
We have combined the SCC algorithm~\cite{spectral_applied} and
kernels to suggest the KSCC algorithm (Algorithm~\ref{alg:kscc}) for
segmenting parametric surfaces which can be mapped to flats in
spaces of moderate dimensions. The computational task is performed
solely in the original data space (using the kernel matrix), thus one would expect KSCC to be faster than performing SCC in the embedded spaces (when having large dimensions). We have
exemplified its success on a few artificial instances of
multi-manifold modeling and on a real-world application of two-view
motion segmentation under perspective camera projection.



There are several important issues that need to be further explored
in order to more broadly and successfully apply the KSCC algorithm.

\begin{figure}
\centering
\includegraphics[width=.23\textwidth]{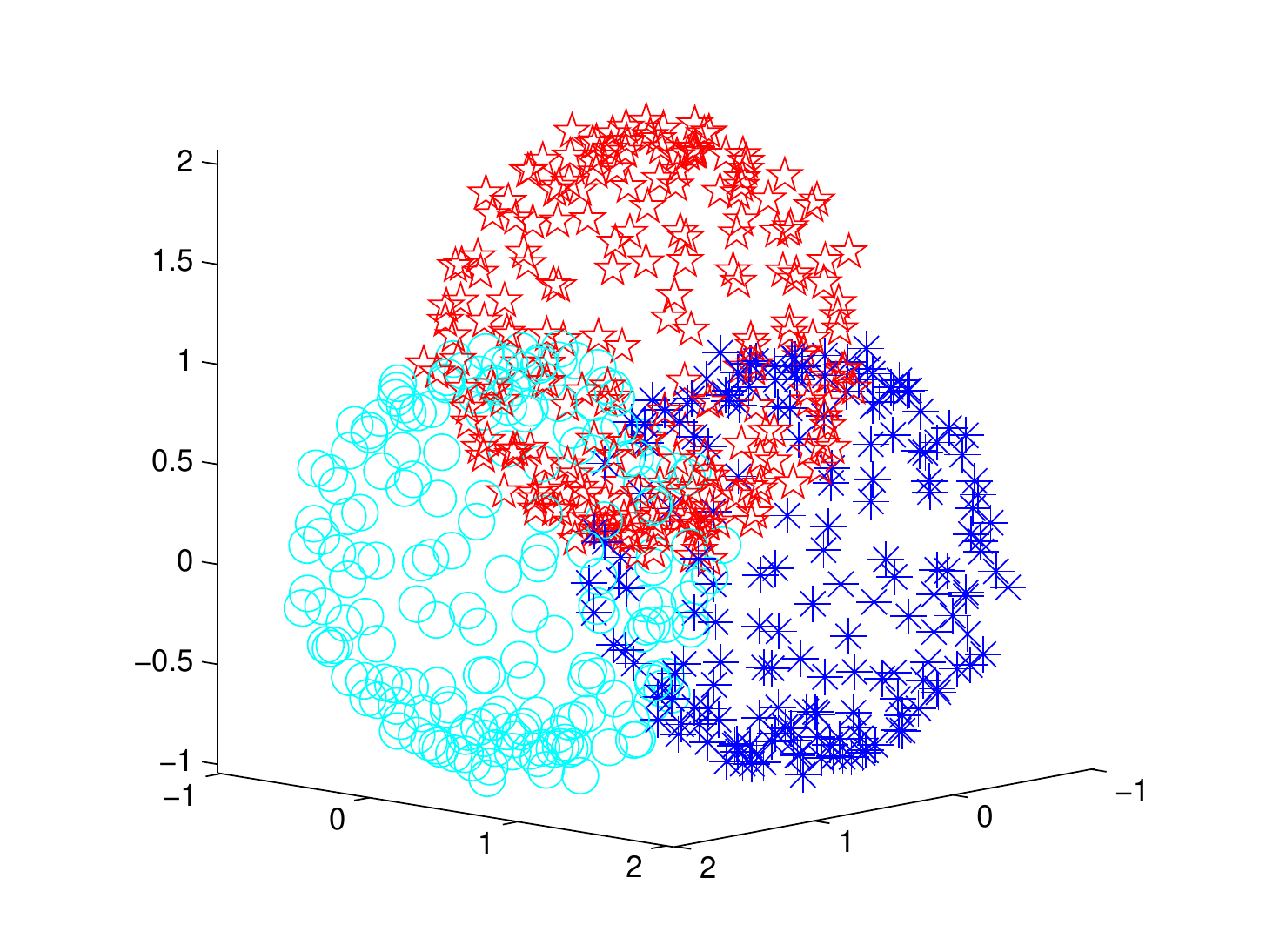}
\includegraphics[width=.23\textwidth]{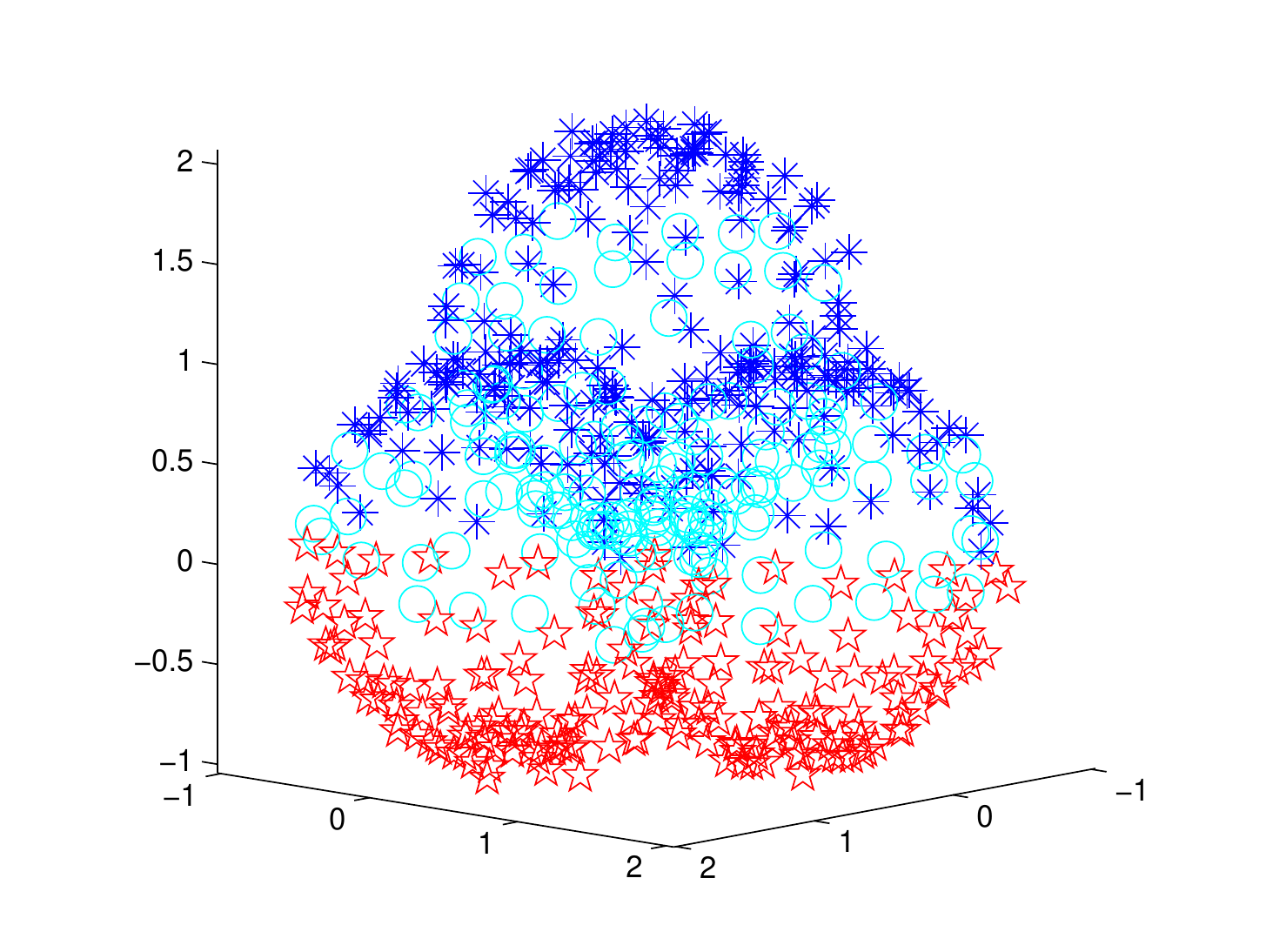}
\caption{Output of the KSCC algorithm when applied to the three
spheres in Fig.~\ref{fig:threespheres_4} with the two kernels
$k_\textrm{2s}$ and $k_\textrm{2f}$, respectively.}
\label{fig:three_spheres_other_kernels}
\end{figure}

1) The choice of the kernel function might affect the success of the
KSCC algorithm. We exemplify this on the three-sphere data in
Fig.~\ref{fig:threespheres_4}, where we used the spherical kernel
(see Eq.~\eqref{eq:spherical_kernel}) and practically segmented
$3$-flats in $\mathbb{R}^4$. We now apply KSCC to the same data with
the following two choices of kernels: the standard quadratic
polynomial kernel
\begin{align}
k_\textrm{2s}(\mathbf{x},\mathbf{y}) = \langle
\mathbf{x},\mathbf{y}\rangle + \langle
\mathbf{x}.^2,\mathbf{y}.^2\rangle,
\end{align}
where $.^2$ means taking elementwise squares, and the full quadratic
polynomial kernel $k_\textrm{2f}(\cdot,\cdot)$ of
Eq.~\eqref{eq:poly2_kernel}, respectively. This is equivalent to
applying SCC to segment $5$-flats in $\mathbb{R}^6$ and $8$-flats in
$\mathbb{R}^9$, respectively. The corresponding results are shown in
Fig.~\ref{fig:three_spheres_other_kernels}.

As it turned out, the KSCC algorithm failed with the full quadratic
kernel $k_\textrm{2f}$. The reason for this is that as $\ell$
increases, the segmentation task becomes more difficult for KSCC
since the initial approximation of the weight matrix $\mathbf{W}$
(defined in Eq.~\eqref{eq:pairwise_weights}) would deteriorate. This
experiment suggests that one should use the optimal kernel function
with KSCC in the sense that it should minimize the intrinsic
dimension of the flats in the feature space. We are currently
developing an automatic scheme to choose the least number of terms
that are necessary for linearization of the manifolds.

2) The dimension of the flats in the feature space is often quite
large, sometimes even with the optimal kernel function (for example,
$\ell = 25$ in the problem of segmenting motions from three
perspective camera views~\cite{Hartley04Three-view}). Due to the
limitation of the KSCC algorithm in dealing with large $\ell$, a
better initialization strategy needs to be explored in order to more
robustly estimate the initial weights $\mathbf{W}$. We plan to
develop such a technique in later research and consequently apply
the improved KSCC algorithm to solve the three-view motion
segmentation problem~\cite{Hartley04Three-view}.


3) We need to examine more carefully the situation when data is
corrupted with noise. Though KSCC can handle small levels of noise,
the clustering task becomes very challenging for KSCC (and probably
for any other manifold clustering algorithm) when the noise level
increases in the original space. There are two reasons for this.
First, in many cases the manifold structure obscures quickly when
corrupted with noise (see e.g., Figs.~\ref{fig:lines_circles} and
\ref{fig:conics_1d}). Second, the noise level is further enlarged in
feature space due to the embedding having higher-order terms. One
has to develop theoretical guarantees for good performance of KSCC
in the presence of noise (as done for SCC
in~\cite{spectral_theory}), and
use related insights for improving the current algorithm. 
For example, one can note the effect of special geometric
transformations of the data, under which the noise distortion (from
original to feature space) is minimal, and apply them before the
KSCC algorithm.

4) We need to study the performance of KSCC on data contaminated
with outliers. Solutions can follow the idea used in the SCC
algorithm~\cite{spectral_applied} and possibly combined with RANdom
SAmple Consensus
(RANSAC)~\cite{Fischler81RANSAC,Torr98geometricmotion,Yang06Robust},
in a similar way as the RAS algorithm~\cite{Rao08RAS}. Future work
will test the KSCC algorithm on the 13 data sets (in
Table~\ref{tab:comparison_2view}) in the presence of outliers.

\section*{Acknowledgements}
We thank the anonymous reviewers for their helpful comments. Special thanks go to Shankar Rao and Yi Ma for extensive email correspondence and help with the RAS algorithm. Thanks to the Institute for Mathematics and its Applications (IMA), in particular Doug Arnold and Fadil Santosa, for an effective hot-topics workshop on
multi-manifold modeling that we all participated in. The research described in this paper was supported by NSF grants \#0612608 and \#0915064.

{\small
\bibliographystyle{ieee}

\begin{thebibliography}{10}\itemsep=-1pt

\bibitem{Agarwal05}
S.~Agarwal, J.~Lim, L.~Zelnik-Manor, P.~Perona, D.~Kriegman, and S.~Belongie.
\newblock Beyond pairwise clustering.
\newblock In {\em Proceedings of the 2005 IEEE Computer Society Conference on
  Computer Vision and Pattern Recognition (CVPR'05)}, volume~2, pages 838--845,
  2005.

\bibitem{Arias-Castro09Spectral}
E.~Arias-Castro, G.~Chen, and G.~Lerman.
\newblock {\em Spectral Clusteirng based on Local Polynomial Approximations}.
\newblock In preparation.

\bibitem{Bradley00kplanes}
P.~Bradley and O.~Mangasarian.
\newblock k-plane clustering.
\newblock {\em J. Global optim.}, 16(1):23--32, 2000.

\bibitem{spectral_theory}
G.~Chen and G.~Lerman.
\newblock Foundations of a multi-way spectral clustering framework for hybrid
  linear modeling.
\newblock {\em Found. Computat. Math. (online)}, 2009.
\newblock DOI 10.1007/s10208-009-9043-7.

\bibitem{spectral_applied}
G.~Chen and G.~Lerman.
\newblock Spectral curvature clustering ({SCC}).
\newblock {\em Int. J. Comput. Vision}, 81(3):317--330, 2009.

\bibitem{Fischler81RANSAC}
M.~Fischler and R.~Bolles.
\newblock Random sample consensus: A paradigm for model fitting with
  applications to image analysis and automated cartography.
\newblock {\em Comm. of the ACM}, 24(6):381--395, June 1981.

\bibitem{Goh08Reimannian}
A.~Goh and R.~Vidal.
\newblock Clustering and dimensionality reduction on {R}iemannian manifolds.
\newblock In {\em CVPR}, pages 1--7, Anchorage, AK, 2008.

\bibitem{Goldberg09ssl}
A.~Goldberg, X.~Zhu, A.~Singh, Z.~Xu, and R.~Nowak.
\newblock Multi-manifold semi-supervised learning.
\newblock In {\em Twelfth International Conference on Artificial Intelligence
  and Statistics (AISTATS)}, 2009.

\bibitem{Haro08TPMM}
G.~Haro, G.~Randall, and G.~Sapiro.
\newblock Translated {P}oisson mixture model for stratification learning.
\newblock {\em Int. J. Comput. Vision}, 80(3):358--374, 2008.

\bibitem{Hartley04Three-view}
R.~Hartley and R.~Vidal.
\newblock The multibody trifocal tensor: Motion segmentation from 3 perspective
  views.
\newblock {\em IEEE Computer Society Conference on Computer Vision and Pattern
  Recognition}.

\bibitem{Ho03}
J.~Ho, M.~Yang, J.~Lim, K.~Lee, and D.~Kriegman.
\newblock Clustering appearances of objects under varying illumination
  conditions.
\newblock In {\em Proceedings of International Conference on Computer Vision
  and Pattern Recognition}, volume~1, pages 11--18, 2003.

\bibitem{Kambhatla94fastnon-linear}
A.~Kambhatla and T.~Leen.
\newblock Fast non-linear dimension reduction.
\newblock In {\em Advances in Neural Information Processing Systems 6}, pages
  152--159, 1994.

\bibitem{Kushnir06multiscale}
D.~Kushnir, M.~Galun, and A.~Brandt.
\newblock Fast multiscale clustering and manifold identification.
\newblock {\em Pattern Recognition}, 39(10):1876--1891, October 2006.

\bibitem{LeCunMNIST}
Y.~LeCun and C.~Cortes.
\newblock The {MNIST} database of handwritten digits.
\newblock http://yann.lecun.com/exdb/mnist/.

\bibitem{Ma07Compression}
Y.~Ma, H.~Derksen, W.~Hong, and J.~Wright.
\newblock Segmentation of multivariate mixed data via lossy coding and
  compression.
\newblock {\em IEEE Transactions on Pattern Analysis and Machine Intelligence},
  29(9):1546--1562, September 2007.

\bibitem{Ma07}
Y.~Ma, A.~Y. Yang, H.~Derksen, and R.~Fossum.
\newblock Estimation of subspace arrangements with applications in modeling and
  segmenting mixed data.
\newblock {\em SIAM Review}, 50(3):413--458, 2008.

\bibitem{Ng02}
A.~Ng, M.~Jordan, and Y.~Weiss.
\newblock On spectral clustering: Analysis and an algorithm.
\newblock In {\em Advances in Neural Information Processing Systems 14}, pages
  849--856, 2001.

\bibitem{Rao08RAS}
S.~Rao, A.~Yang, S.~Sastry, and Y.~Ma.
\newblock Robust algebraic segmentation of mixed rigid-body and planar motions.
\newblock Available at
  http://www.eecs.berkeley.edu/$\sim$yang/paper/IJCV08-RAS.pdf, 2008.

\bibitem{Scholkopf02Kernels}
B.~Sch\"{o}lkopf and A.~J. Smola.
\newblock {\em Learning with Kernels}.
\newblock The MIT Press, Cambridge, Massachusetts, 2002.

\bibitem{Souvenir05}
R.~Souvenir and R.~Pless.
\newblock Manifold clustering.
\newblock In {\em the 10th International Conference on Computer Vision (ICCV
  2005)}, 2005.

\bibitem{Tipping99mixtures}
M.~Tipping and C.~Bishop.
\newblock Mixtures of probabilistic principal component analysers.
\newblock {\em Neural Computation}, 11(2):443--482, 1999.

\bibitem{Torr98geometricmotion}
P.~H.~S. Torr.
\newblock Geometric motion segmentation and model selection.
\newblock {\em Phil. Trans. R. Soc. Lond. A}, 356:1321--1340, 1998.

\bibitem{Tseng00nearest}
P.~Tseng.
\newblock Nearest $q$-flat to $m$ points.
\newblock {\em Journal of Optimization Theory and Applications},
  105(1):249--252, April 2000.

\bibitem{Vidal05}
R.~Vidal, Y.~Ma, and S.~Sastry.
\newblock Generalized principal component analysis ({GPCA}).
\newblock {\em IEEE Transactions on Pattern Analysis and Machine Intelligence},
  27(12), 2005.

\bibitem{Yan06LSA}
J.~Yan and M.~Pollefeys.
\newblock A general framework for motion segmentation: Independent,
  articulated, rigid, non-rigid, degenerate and nondegenerate.
\newblock In {\em ECCV}, volume~4, pages 94--106, 2006.

\bibitem{Yang06Robust}
A.~Y. Yang, S.~R. Rao, and Y.~Ma.
\newblock Robust statistical estimation and segmentation of multiple subspaces.
\newblock In {\em Computer Vision and Pattern Recognition Workshop}, June 2006.

\end{thebibliography}

}

\end{document}